\documentclass{article}
\usepackage{rotating}
\usepackage{algorithm}
\usepackage{algpseudocode}
\usepackage{arxiv}
\usepackage{multirow}
\usepackage[utf8]{inputenc} 
\usepackage[T1]{fontenc}    
\usepackage{hyperref}       
\usepackage{url}            
\usepackage{booktabs}       
\usepackage{amsfonts}       
\usepackage{nicefrac}       
\usepackage{microtype}      
\usepackage{lipsum}
\usepackage{graphicx}
\usepackage{xcolor}
\graphicspath{ {./images/} }
\usepackage{subfigure}
\date{}

\usepackage{amsmath}
\usepackage{amssymb}
\usepackage{bm}

\title{FedMHAR: Federated Multimodal Human Activity Recognition using Multi-Agent Reinforcement Learning}

\author{
 Debasmita Dey \\
  SQC \& OR Unit\\
Indian Statistical Institute Kolkata \\
Kolkata, India \\
  \texttt{debasmita.dey9\_r@isical.ac.in} \\
   \And
Tanmay Sen \\
SQC \& OR Unit\\
Indian Statistical Institute Kolkata \\
Kolkata, India \\
\thanks{Corresponding author: tanmay.sen@isical.ac.in} \texttt{tanmay.sen@isical.ac.in}\\
  \And
Himel Mallick \\
  Division of Gastroenterology and Hepatology, Weill Cornell Medicine, New York, NY, USA\\
  Department of Population Health Sciences, Weill Cornell Medicine, New York, NY, USA \\
  Department of Statistics and Data Science, Cornell University, Ithaca, NY, USA\\
}

\algnewcommand\algorithmicinput{\textbf{Input:}}
\algnewcommand\algorithmicoutput{\textbf{Output:}}
\algnewcommand\INPUT{\item[\algorithmicinput]}
\algnewcommand\OUTPUT{\item[\algorithmicoutput]}
\algnewcommand\HYPERPARAM{\item[\textbf{Hyperparameters:}]}

\begin{document}

\maketitle

\begin{abstract}

Human Activity Recognition (HAR) from heterogeneous  wearable sensors has become a fundamental component of the Internet of Health Things (IoHT), enabling continuous monitoring for rehabilitation, elderly care, and smart healthcare applications. Yet most existing multimodal fusion methods assign fixed equal weights to all sensor streams, overlooking the fact that different modalities carry different information depending on the subject and activity, and that some sensors are far more expensive to acquire than others. The quality of each sensor changes over time because of movement, incorrect placement, or temporary blockage. Therefore, assigning fixed importance to all sensor modalities is often ineffective in real world applications.  In this paper, we first propose an adaptive and cost aware multimodal HAR framework based on multi-agent reinforcement learning for the centralized setting. We then extend this framework to federated learning, resulting in the proposed \textit{FedMHAR} framework. 
In the centralized setting, we model multimodal fusion as a cooperative Multi-Agent Reinforcement Learning (MARL) problem, where each sensing modality is assigned a PPO based agent that learns a per sample fusion weight for its corresponding sensor, enabling the model to dynamically emphasize informative modalities while down weighting costly sensors whenever cheaper alternatives provide sufficient information. 
In the federated setting, we introduce BiFL-PPO, a novel bidirectional federated optimization strategy in which a server side PPO policy learns client specific trust weights that are fed back to clients to adaptively adjust local learning rates and proximal regularization. Unlike conventional round level optimization, BiFL-PPO uses dense batch level rewards to provide more frequent feedback, leading to more stable training under heterogeneous client data. We evaluate the proposed framework on the MEx Rehabilitation and UTD Multimodal Human Action datasets. In the centralized setting, the proposed MARL based framework achieves classification accuracies of 87.30\% and 94.98\%, outperforming both conventional multimodal fusion methods and state-of-the-art HAR models. In the federated setting, \textit{FedMHAR} achieves 79.74\% and 77.49\% accuracy, consistently surpassing strong federated baselines, including FedAvg, FedProx, FedBN, FedNova, and AdaFedProx. Across both datasets, the proposed bidirectional PPO strategy yields more stable performance than standard federated optimization while reducing sensor acquisition cost.

\end{abstract}

\section{Introduction}

Human Activity Recognition (HAR) \cite{ahmad2021graph, sunny2015applications, yang2024cross, chatzaki2016human, wan2020deep} has emerged as an important research area in computer vision, wearable computing, and smart healthcare due to its broad range of applications, including rehabilitation monitoring, elderly care, patient surveillance, and assisted living. Recent advances in wearable devices, ambient sensing technologies, and edge computing have enabled continuous monitoring of human activities through rich streams of sensor data, making HAR an essential component of intelligent healthcare and smart environments. Accurate recognition of human activities not only improves personalized healthcare services but also enables timely intervention in safety-critical situations. To improve recognition performance, modern HAR systems increasingly exploit multiple sensing modalities, including accelerometers, gyroscopes, pressure sensors, depth cameras, skeletal joints, and electromyography (EMG) signals \cite{de2015multimodal, khan2025multimodal, bandyopadhyay2025mharfedllm}. These heterogeneous modalities capture complementary characteristics of human motion and environmental interactions that cannot be adequately represented by a single sensor. Multimodal HAR thus falls under the broader umbrella of multimodal artificial intelligence (AI), in which models learn jointly from heterogeneous data sources, and has consistently demonstrated superior recognition accuracy and robustness compared with unimodal approaches, particularly for complex activities involving subtle body movements or occlusions, mirroring its success across diverse biomedical domains \cite{mallick2024integrated, roy2025bayesian, anceschi2026bayesian, roy2026conformalized}.

Motivated by these observations, we propose a cooperative Multi-Agent Reinforcement Learning framework for adaptive multimodal fusion. Specifically, each sensing modality is associated with an independent Proximal Policy Optimization (PPO) agent that learns a per-sample fusion weight reflecting the usefulness of its corresponding sensor. The learned fusion weights are employed by a Mixture-of-Experts (MoE) \cite{cai2024survey} architecture consisting of modality-specific expert networks, allowing the model to dynamically combine multimodal representations according to the current activity instance. By explicitly incorporating normalized sensor acquisition cost into the reinforcement learning reward, the proposed framework performs implicit sensor scheduling, automatically down-weighting expensive modalities whenever lower-cost sensors provide sufficient discriminative information. Unlike conventional graph- or attention-based fusion methods, the proposed policy adapts continuously during inference, making it substantially more robust to changing sensing conditions.

Reinforcement Learning (RL) \cite{sutton1998reinforcement} offers a natural solution to these challenges by formulating multimodal fusion as a sequential decision-making problem. Rather than learning fixed fusion weights, an RL agent in general interacts with the environment and learns adaptive policies that maximize long-term reward. Such a reward can encourage high recognition accuracy while penalizing the use of expensive sensing modalities. Consequently, RL enables dynamic, per-sample sensor weighting and adaptive fusion according to the current sensing conditions, allowing the recognition model to emphasize informative modalities while suppressing unreliable or unnecessarily costly sensors. This capability makes RL particularly attractive for real-world HAR systems operating in dynamic environments and under resource constraints.

Motivated by these observations, we propose a cooperative Multi-Agent Reinforcement Learning framework for adaptive multimodal fusion. Specifically, each sensing modality is associated with an independent Proximal Policy Optimization (PPO) agent that learns a per-sample fusion weight reflecting the usefulness of its corresponding sensor. The learned fusion weights are employed by a Mixture-of-Experts (MoE) \cite{cai2024survey} architecture consisting of modality-specific expert networks, allowing the model to dynamically combine multimodal representations according to the current activity instance. By explicitly incorporating normalized sensor acquisition cost into the reinforcement learning reward, the proposed framework performs implicit sensor scheduling, automatically down-weighting expensive modalities whenever lower-cost sensors provide sufficient discriminative information. Unlike conventional graph- or attention-based fusion methods, the proposed policy adapts continuously during inference, making it substantially more robust to changing sensing conditions.

Although centralized training has demonstrated remarkable success for HAR, collecting wearable sensor data in a centralized repository raises significant privacy and security concerns, particularly in healthcare applications where activity records often contain highly sensitive personal information. Federated Learning (FL) addresses this challenge by enabling multiple devices or institutions to collaboratively train a shared model without exchanging raw data \cite{mcmahan2017communication, wang2025privacy}. However, wearable HAR datasets are inherently heterogeneous because activity patterns, sensor quality, and user behavior vary considerably across clients. Conventional federated optimization algorithms such as FedAvg \cite{mcmahan2017communication} often struggle under such non-IID data distributions, leading to unstable convergence and degraded recognition performance \cite{zhao2018federated, li2020convergence, karimireddy2020scaffold, li2020federated}. To address these limitations, we extend the proposed centralized framework to the federated setting through \textit{FedMHAR}, a privacy-preserving multimodal HAR framework. At its core, \textit{FedMHAR} introduces BiFL-PPO, a bidirectional federated optimization strategy in which a server-side PPO policy learns client-specific trust weights based on observed training behavior. These trust weights are used both to weight client updates during aggregation and, when communicated back to participating clients, to adaptively adjust local learning rates and proximal regularization during training, allowing the optimization process to account for heterogeneous client characteristics. By coupling server-side aggregation with client-side optimization through a shared trust signal, BiFL-PPO yields more stable optimization under heterogeneous data distributions while preserving user privacy.

The main contributions of this work are summarized as follows:
\begin{itemize}
    \item We propose a cooperative multi-agent PPO (MAPPO) framework for adaptive multimodal fusion in HAR, in which each sensing modality is assigned an agent that learns per-sample fusion weights for an MoE of modality-specific expert networks, according to the reliability and informativeness of its sensor.

    \item We introduce a cost-aware reward formulation that explicitly incorporates sensor acquisition cost, encouraging reliance on lower-cost modalities without requiring supervision on sensor quality.

    \item We extend the proposed approach to federated learning through \textit{FedMHAR} with BiFL-PPO, a bidirectional PPO-based optimization strategy in which learned client-specific trust weights guide both server-side aggregation and client-side local optimization under heterogeneous client data.

    \item Experiments on the MEx Rehabilitation \cite{wijekoon2019MEx} and UTD Multimodal Human Action \cite{7350781} datasets show that the proposed framework outperforms equal-weight fusion and existing centralized HAR methods, and improves on standard federated baselines with more stable performance across seeds.
\end{itemize}












\section{Related Work}
This section reviews the literature on Human Activity Recognition (HAR), Deep Reinforcement Learning (DRL), Federated Learning (FL), and Mixture of Experts (MoE).
\subsection{Human Activity Recognition}
Human Activity Recognition (HAR) has gained significant attention due to its applications in healthcare, smart homes, surveillance, rehabilitation, and fitness monitoring \cite{kumar2024human}. Early HAR systems relied on traditional machine learning techniques with handcrafted features, which often struggled with complex temporal patterns and real-world variability \cite{gupta2022human}. Recent advancements in deep learning, particularly CNNs, LSTMs, RNNs, Transformers, and hybrid architectures, have significantly improved activity recognition accuracy by automatically learning spatial and temporal features from raw sensor data collected through smartphones, wearables, and IoT devices \cite{zhou2025efficient}. Recent studies also emphasize multimodal sensor fusion, attention mechanisms, federated learning, and edge computing to enhance robustness, privacy, and real-time performance. In this direction, Islam et al. \cite{islam2022multimodal} combine CNN and ConvLSTM features through a self-attention mechanism for multimodal HAR, demonstrating that fusing visual and time-series sensor data yields more robust activity recognition than single-modality approaches. Building on this, Wang et al. \cite{wang2026multimodal} propose M2HAR-Net, fusing Wi-Fi CSI signals with smartwatch inertial data through parallel CNN-Transformer encoders, showing that combining wireless sensing with wearable sensors captures complementary motion characteristics across heterogeneous modalities. However, challenges such as sensor noise, data imbalance, computational overhead, and poor cross-domain generalization \cite{javadi2025graph} still remain active research issues.

\subsection{Deep Reinforcement Learning in Human Activity Recognition}
Deep reinforcement learning has been applied to human activity recognition across several distinct problem formulations \cite{zhou2020deep}. Tang et al. \cite{tang2018deep} introduced a deep progressive reinforcement learning (DPRL) framework in which an agent sequentially identifies the most informative frames from a skeleton-based video sequence. Wu et al. \cite{wu2019multi} formulated the frame selection problem as a set of multiple Markov Decision Processes (MDPs), with each agent assigned the task of selecting a specific frame. In \cite{dong2019attention}, an attention-aware sampling approach was proposed to preserve only the most informative activity frames and discard irrelevant video frames. In \cite{xu2020adaptive}, a Feature Selection Network (FSN) based on the Actor-Critic RL framework was proposed to select discriminative frames from skeleton sequences. The frame representations are obtained through a Generalized Graph Convolutional Network (GGCN).
While prior DRL-based HAR methods focus on single-modality frame selection using a single agent with no cost awareness or federated capability, our MAPPO-MoE assigns cooperative multi-agent policies across heterogeneous wearable sensor streams with explicit sensor cost penalisation, per-sample dynamic fusion, and a federated extension under non-IID data partitioning.

\subsection{Federated Learning}Federated learning was introduced by McMahan~\textit{et al.}~\cite{mcmahan2017communication}, who proposed FedAvg, where clients perform local SGD and the server aggregates updates through weighted averaging. Although effective, FedAvg struggles under statistical and systems heterogeneity. To address client drift, Li~\textit{et al.}~\cite{li2020federated} proposed FedProx, which adds a proximal regularization term to the local objective, improving convergence under non-IID data. Wang~\textit{et al.}~\cite{wang2020tackling} introduced FedNova, which normalizes client updates to eliminate aggregation bias caused by varying local update counts. Li~\textit{et al.}~\cite{li2021fedbn} proposed FedBN, which keeps batch-normalization layers local to mitigate feature-distribution shift across clients. More recently, Sahoo~\textit{et al.}~\cite{sahoo2024adafedprox} introduced AdaFedProx, where a server-side DQN adaptively learns the proximal coefficient based on client states, improving performance over FedAvg, FedProx, FedNova, and FedBN under heterogeneous settings.

Unlike these approaches, our method learns aggregation weights directly using a Server-PPO agent. At each communication round, the agent observes per-client validation signals and generates adaptive aggregation weights through a Beta policy, optimizing global model performance instead of relying on fixed aggregation rules or manually designed regularization.

\subsection{Mixture of Experts (MoE)}
Recent advances have demonstrated the effectiveness of MoE architectures across a wide range of domains by enabling dynamic expert specialization and efficient computation. The authors in \cite{lin2026moe} integrated MoE-based large language model, where a router dynamically activates the top-k experts to process multimodal vision-language tokens, improving reasoning. Similarly, for dynamically routing visual features to domain-specialized experts, the authors in \cite{gao2026more} have instilled MoE. MoE also found its use case in solving heterogeneity problems in PDE datasets \cite{wang2023mhagnn}. For vision language action models in end-to-end autonomous driving \cite{yang2026drivemoe}, MoE has played a vital role in perception, reasoning, and action generation. Using LoRA ranks as an expert of the MoE has helped in solving multi-task model merging, reducing parameter interference. Building on these ideas, our work also adopts the MoE paradigm for multimodal fusion, augmented with a Proximal Policy Optimization (PPO) agent that learns modality-expert weighting as a sequential decision-making problem. This design enables dynamic, sample-level fusion rather than static concatenation, allowing the model to lean more heavily on whichever modality is most informative for a given instance, while an accuracy- and cost-aware reward further encourages the policy to favor cheaper or more readily available sensors over costlier ones when they are sufficient for correct classification.

\section{Background Concepts}

\subsection{Multi-Agent Proximal Policy Optimization}

Multi-Agent Proximal Policy Optimization (MAPPO) is a policy gradient algorithm designed for cooperative multi-agent reinforcement learning (MARL). It extends the Proximal Policy Optimization (PPO) \cite{schulman2017proximal} algorithm to multi-agent settings while following the \emph{Centralized Training and Decentralized Execution (CTDE)} paradigm \cite{yu2022surprising}. During training, agents utilize global state information through a centralized critic to stabilize learning, whereas during execution each agent independently selects actions using only its local observations.

Consider a cooperative  game defined by the tuple \(\mathcal{G}=\left(\mathcal{S},\{\mathcal{O}_i\}_{i=1}^{N},
\{\mathcal{A}_i\}_{i=1}^{N},
P,R,\gamma\right),
\) where $\mathcal{S}$ denotes the global state space, $\mathcal{O}_i$ and $\mathcal{A}_i$ represent the observation and action spaces of agent $i$, respectively, $P(s'|s,\mathbf{a})$ is the state transition probability, $R(s,\mathbf{a})$ is the shared team reward function, $\gamma\in[0,1)$ is the discount factor, and $N$ is the total number of agents. At each time step $t$, the environment is in state $s_t$. Agent $i$ receives a local observation $o_i^t\in\mathcal{O}_i$ and samples an action according to its stochastic policy $_i^t \sim \pi_{\theta}(a_i^t|o_i^t)$,
where $\pi_{\theta}$ denotes the policy network parameterized by $\theta$. Although each agent acts independently using its own observation, MAPPO commonly employs parameter sharing, where all agents use the same policy network while conditioning on different observations. The environment then transitions to the next state according to $s_{t+1}\sim P(s_{t+1}|s_t,\mathbf{a}_t)$, where $\mathbf{a}_t=(a_1^t,a_2^t,\ldots,a_N^t)$ represents the joint action of all agents. Since the environment is cooperative, all agents receive the same team reward $r_t$. The objective of MAPPO is to maximize the expected cumulative discounted team reward

\begin{equation}
J(\theta)=
\mathbb{E}_{\tau\sim\pi_\theta}
\left[
\sum_{t=0}^{T}
\gamma^t r_t
\right],
\end{equation}

where $\tau$ denotes a trajectory generated by the joint policy and $T$ is the episode horizon.

\paragraph{Centralized Critic: } During training, MAPPO employs a centralized value function that utilizes the global state to estimate the expected return $V_{\phi}(s_t),$
where $\phi$ denotes the critic parameters. The centralized critic has access to complete state information only during training, while the actor uses only local observations during execution, thereby satisfying the CTDE paradigm.

\paragraph{Generalized Advantage Estimation: }

To reduce the variance of policy gradient estimates while maintaining low bias, MAPPO computes the advantage function using Generalized Advantage Estimation (GAE)

\begin{equation}
A_t=
\sum_{l=0}^{T-t-1}
(\gamma\lambda)^l
\delta_{t+l},
\end{equation}

where $\lambda\in[0,1]$ is the GAE smoothing parameter and

\begin{equation*}
\delta_t=
r_t+
\gamma V_{\phi}(s_{t+1})
-
V_{\phi}(s_t)
\end{equation*}

is the temporal-difference (TD) error. The estimated return is subsequently computed as

\begin{equation*}
R_t=A_t+V_{\phi}(s_t).
\end{equation*}

\paragraph{Policy Optimization: }

The actor network is updated using the PPO clipped surrogate objective. The probability ratio between the updated policy and the previous policy is defined by \cite{schulman2017proximal}
 
\begin{equation*}
r_t(\theta)=
\frac{\pi_{\theta}(a_i^t|o_i^t)}
{\pi_{\theta_{\mathrm{old}}}(a_i^t|o_i^t)}.
\end{equation*}

The clipped objective is

\begin{equation*}
L^{\mathrm{CLIP}}(\theta)
=
\mathbb{E}
\left[
\min
\left(
r_t(\theta)A_t,
\operatorname{clip}
\left(
r_t(\theta),
1-\epsilon,
1+\epsilon
\right)
A_t
\right)
\right],
\end{equation*}

where $\epsilon$ is the clipping parameter controlling the maximum policy update. The clipping operation prevents excessively large policy updates, thereby improving training stability.

\paragraph{Value Function Loss: }

The critic is optimized by minimizing the squared error between the predicted value and the estimated return
\[
L_V(\phi)
=
\mathbb{E}
\left[
\left(
V_{\phi}(s_t)-R_t
\right)^2
\right].
\]

\paragraph{Entropy Regularization: }

To encourage exploration and avoid premature convergence, MAPPO incorporates an entropy regularization term
\(
L_{\mathrm{ent}}
=
\mathbb{E}
\left[
H(\pi_{\theta}(\cdot|o_i))
\right],
\)
where $H(\cdot)$ denotes the Shannon entropy of the policy distribution.

Finally, the  MAPPO optimization objective is

\begin{equation*}
L_{\mathrm{MAPPO}}
=
L^{\mathrm{CLIP}}
-
c_1L_V
+
c_2L_{\mathrm{ent}},
\end{equation*}

where $c_1$ and $c_2$ are weighting coefficients that balance policy optimization, value estimation, and exploration. This objective enables stable policy learning while effectively coordinating multiple agents in cooperative environments. 











To improve specialization across diverse input patterns, Mixture of Experts (MoE) \cite{jacobs1991adaptive} architectures have recently gained significant attention. Instead of learning a single shared representation, MoE enables multiple expert networks to specialize in different regions of the input space while a gating network dynamically determines their contributions. Therefore, the basic formulation of MoE is briefly described next.

\subsection{Mixture of Experts}

Mixture of Experts (MoE) \cite{jacobs1991adaptive} is a neural network architecture that improves learning efficiency and model expressiveness by decomposing a complex learning task into multiple specialized expert networks. Rather than relying on a single network to capture all data characteristics, MoE employs several experts, each of which learns complementary representations of the input. A gating network then adaptively combines the outputs of these experts based on the input sample.

Consider an input feature vector $x \in \mathbb{R}^{d}$. An MoE model consists of a set of $N$ expert networks,
\(
\mathcal{E}=\{E_1,E_2,\ldots,E_N\},
\)
where each expert independently processes the input and produces an output
\(
y_i=E_i(x), \qquad i=1,\ldots,N.
\)
To determine the contribution of each expert, a gating network $G(\cdot)$ computes a score vector $z = G(x)$, where $z=[z_1,z_2,\ldots,z_N]$ contains the unnormalized gating scores. These scores are converted into normalized gating weights using the softmax function,
\(
g_i(x)=
\frac{\exp(z_i)}
{\sum_{j=1}^{N}\exp(z_j)},
\)
where
\(
\sum_{i=1}^{N} g_i(x)=1,
\qquad
g_i(x)\ge0.
\)
The final MoE output is obtained as the weighted combination of all expert predictions,
\(
y=
\sum_{i=1}^{N}
g_i(x)\,E_i(x)
=
\sum_{i=1}^{N}
g_i(x)\,y_i.
\)

The adaptive weighting mechanism enables different experts to specialize in distinct input characteristics while allowing the gating network to dynamically select the most relevant experts for each input sample. Consequently, MoE improves the representational capacity and flexibility of deep neural networks without requiring all experts to contribute equally to every prediction.











\subsection{Federated Learning}

Federated Learning (FL) \cite{mcmahan2017communication} is a distributed machine learning approach where multiple clients collaboratively train a global model without sharing their raw data, thereby preserving privacy. Suppose there are $K$ clients, each having a local dataset $D_k$ with $n_k$ samples. The global objective function is defined as:

\begin{equation}
  F(w) = \sum_{k=1}^{K} \frac{n_k}{N} F_k(w)  
\end{equation}

where $N = \sum_{k=1}^{K} n_k$, $w$ represents the global model parameters, and $F_k(w)$ is the local loss function of client $k$. At communication round $t$, the server broadcasts the current global model $w_t$ to all participating clients. Each client performs local optimization on its own dataset,

\begin{equation}
w_k^{t+1} = w_t - \eta \nabla F_k(w_t),
\end{equation}

where $\eta$ is the learning rate. After local training, only the updated model parameters are transmitted to the server. The server then aggregates the received models to obtain an updated global model, which is subsequently broadcast to all clients for the next communication round. This iterative optimization process enables collaborative learning while ensuring that the raw training data remain local to each client.




Although the general federated learning framework remains unchanged, several aggregation algorithms have been developed to address challenges such as statistical heterogeneity, client drift, and feature distribution mismatch. Since these algorithms serve as baseline methods for evaluating the proposed framework, their fundamental principles are briefly described in the following subsection.

\subsection{Federated Aggregation Algorithms} \label{fed_aggr}




\paragraph{FedAvg: } FedAvg \cite{mcmahan2017communication} performs local training by allowing each client to optimize its model parameters using standard stochastic gradient descent on local data for multiple epochs. Let $w_t$ denote the global model at communication round $t$, and $w_k$ denote the locally updated model of client $k$. The vanilla FedAvg server aggregates client updates using data-size proportional averaging,
\(
w_{t+1}=\sum_{k=1}^{K}\frac{n_k}{n}w_k,
\)
where $n_k$ is the number of samples held by client $k$ and $n=\sum_{k=1}^{K}n_k$. 


\paragraph{FedProx: } FedProx \cite{li2020federated}  performs local optimization by minimizing the objective
\(
\min_{w_k} F_k(w_k)+\frac{\mu}{2}\|w_k-w_t\|^2,
\)
where \(F_k(w_k)\) is the local loss, \(w_t\) is the current global model, and \(\mu\) is the proximal coefficient that limits client drift under heterogeneous data distributions. After local training, the vanilla FedProx server updates the global model using weighted averaging,
\(
w_{t+1}=\sum_{k=1}^{K}\frac{n_k}{n}w_k.
\)



\paragraph{Fednova: } FedNova \cite{wang2020tackling} mitigates the objective inconsistency caused by heterogeneous local training by normalizing each client's accumulated model update before aggregation. Specifically, the server updates the global model as
\(
w_{t+1}=w_t-\tau_{\mathrm{eff}}\sum_{k=1}^{K}p_k d_k,
\)
where \(w_t\) and \(w_{t+1}\) denote the global model parameters before and after communication round \(t\), \(K\) is the total number of clients, \(p_k=n_k/n\) is the data-size weight of client \(k\) with \(n_k\) local samples out of \(n\) total samples, \(d_k=\Delta_k/\tau_k\) is the normalized local update, \(\Delta_k=w_k-w_t\) is the accumulated model update after local training, \(\tau_k\) is the number of local optimization steps performed by client \(k\), and \(\tau_{\mathrm{eff}}=\sum_{k=1}^{K}p_k\tau_k\) is the effective weighted average of local updates. 


\paragraph{FedBN: } FedBN \cite{li2021fedbn} mitigates feature heterogeneity by keeping Batch Normalization (BN) parameters local while aggregating only the non-BN parameters. Let the model be represented as $(w=(w^{NB},w^{BN}))$, where $(w^{NB})$ and $(w^{BN})$ denote the non-BN and BN parameters, respectively. Vanilla FedBN aggregates the non-BN parameters as
\(
w_{t+1}^{NB}=\sum_{k=1}^{K}p_k,w_k^{NB},
\)
where (K) is the number of clients, $(w_k^{NB})$ is the locally trained non-BN model of client $(k)$, and $(p_k=n_k/n)$ is the data-size weight. 


\paragraph{AdaFedProx: } AdaFedProx \cite{sahoo2024adafedprox} extends FedProx by performing local optimization with a proximal regularization term,
\(
w_k^{t+1}
=
\arg\min_{w}
\left[
F_k(w)
+
\frac{\mu}{2}\|w-w_t\|^2
\right],
\)
where \(F_k(w)\) denotes the local objective of client \(k\), \(w_t\) is the global model at communication round \(t\), \(\mu\) is the proximal regularization coefficient, and \(w_k^{t+1}\) is the updated local model. AdaFedProx aggregates the client models using fixed or uniform averaging,
\(
w_{t+1}
=
\sum_{k=1}^{K} p_k w_k^{t+1},
\)
where \(p_k\) denotes the aggregation weight of client \(k\).



\section{Proposed Methodology}

This section presents the proposed framework, which consists of two complementary components: (i) an adaptive and cost aware multimodal human activity recognition framework for the centralized setting, and (ii) its extension to privacy-preserving federated learning through the proposed \textbf{BiFL-PPO} optimization strategy. 

In the centralized setting, the proposed framework formulates multimodal fusion as a cooperative multi-agent reinforcement learning problem. Each sensing modality is associated with a dedicated expert classifier and a MAPPO agent, which jointly learn adaptive modality specific trust weights for dynamic prediction fusion. To enable privacy-preserving collaborative learning, the framework is further extended to the federated setting through the proposed \textbf{BiFL-PPO} strategy. A server side PPO agent learns adaptive client trust weights that are used for both model aggregation and feedback guided local optimization, establishing a bidirectional interaction between the server and participating clients. The detailed formulation of both centralized and federated frameworks is presented in the following subsections.


The architectures of the proposed centralized and federated frameworks are illustrated in Figs.~\ref{fig:cen_arch} and \ref{fig:cen_arch_fl}, respectively. The following subsections describe the individual components and their corresponding formulations in detail.


\subsection{Centralised Architecture} \label{proposed_cen}
The proposed architecture is a multimodal, multi-stage framework that integrates deep learning based feature extraction with MARL for adaptive sensor fusion. It is designed to process heterogeneous sensor data and produce accurate activity classification by dynamically adjusting the contribution of each modality.


\begin{figure} 
    \centering
    \includegraphics[scale = 0.45]{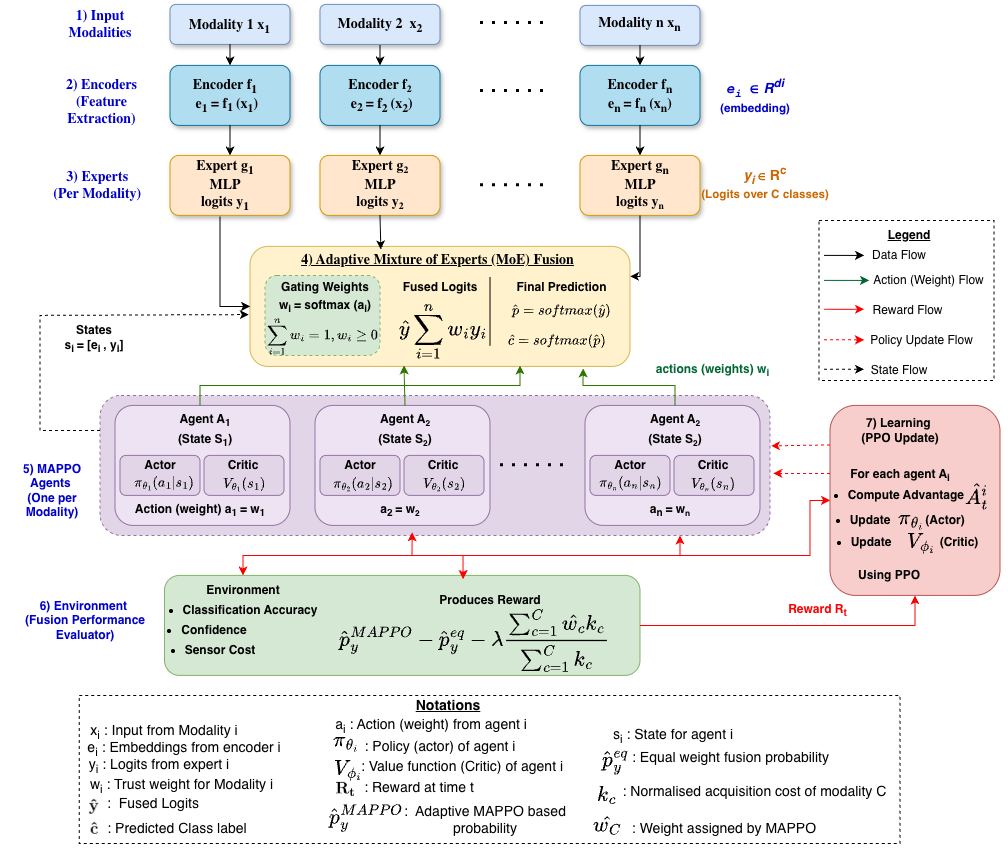}
    \caption{Overview of the centralized MAPPO-MoE architecture. Each modality's embedding, extracted by a shared multimodal encoder, is concatenated into a joint state $s$ observed by every modality's MAPPO agent; each agent outputs a trust weight $w_m$ for its own modality's expert via a Beta policy, and these weights are combined via softmax to fuse the experts' logits into the final prediction. A shared critic $V(s)$ estimates the value of the state, and a reward based on the classification-accuracy gain over an equal-weight baseline, penalized by sensor cost, is used to periodically update the agents and critic via a clipped-surrogate PPO objective.}
    \label{fig:cen_arch}
    \vspace{-3mm}
\end{figure}

\paragraph{Stage 1. Modality Specific Feature Extraction:}

As illustrated in Fig.~\ref{fig:cen_arch}, each sensing modality is processed independently using a dedicated one-dimensional convolutional neural network (1D CNN) encoder. Let $\mathbf{x}_c \in \mathbb{R}^{T_c \times d_c}$ denote the input sequence corresponding to the $c$-th sensing modality, where $T_c$ is the sequence length and $d_c$ is the input feature dimension. The encoder $f_c(\cdot)$ transforms the raw sensor signal into a latent feature representation,

\begin{equation}
\mathbf{e}_c = f_c(\mathbf{x}_c), \qquad \mathbf{e}_c \in \mathbb{R}^{d_e},
\end{equation}

where $d_e$ denotes the embedding dimension. For all experiments reported in this paper, we set $d_e=128$. Since each encoder is trained exclusively on its corresponding sensing modality, it learns modality-specific temporal characteristics while preserving discriminative information for subsequent activity classification. The extracted embeddings serve as the input to the corresponding expert networks in the subsequent stage, where modality-specific predictions are generated prior to adaptive multimodal fusion.




\paragraph{Stage 2.  Expert Level Classification:}

The modality-specific embeddings extracted in Stage~1 are subsequently processed by independent expert networks. Each expert is implemented as a multilayer perceptron (MLP) and is responsible for learning discriminative decision boundaries for its corresponding sensing modality. Given the embedding $\mathbf{e}_c$, the $c$-th expert produces a modality-specific logit vector

\begin{equation}
\mathbf{y}_c
=
g_c(\mathbf{e}_c),
\qquad
c=1,\ldots,C,
\end{equation}

where $g_c(\cdot)$ denotes the expert classifier corresponding to modality $c$, and $\mathbf{y}_c\in\mathbb{R}^{C}$ is the predicted logit vector over the $C$ activity classes.

During the pretraining stage, the expert networks are jointly optimized using cross-entropy loss under equal-weight fusion. Specifically, the initial fused prediction is obtained by uniformly averaging the expert logits,

\begin{equation}
\hat{\mathbf{z}}^{\mathrm{eq}}
=
\frac{1}{C}
\sum_{c=1}^{C}
\mathbf{z}_c,
\end{equation}

where $\hat{\mathbf{z}}^{\mathrm{eq}}$ denotes the equal-weight fused logits. After applying the softmax function, the resulting probability of the ground-truth class, denoted by $\hat{p}^{\mathrm{eq}}_{y}$, serves as the reference confidence used in the reward formulation described in Stage~3.


\paragraph{Stage 3. Adaptive Weight Generation using MAPPO:}

Instead of assigning equal importance to all sensing modalities, the proposed framework formulates multimodal fusion as a cooperative multi-agent reinforcement learning problem. Each sensing modality is associated with an independent MAPPO agent that learns a sample-specific fusion weight based on the global multimodal representation.

At each training step, the shared state is constructed by concatenating the embeddings generated by all modality-specific encoders,
\(
\mathbf{s}
=
\left[
\mathbf{e}_1
\,\|\,\mathbf{e}_2
\,\|\,\cdots
\,\|\,\mathbf{e}_N
\right],
\)

where $\mathbf{e}_n$ denotes the embedding extracted by the $n$-th encoder in Stage~1 and $\|$ represents vector concatenation. During centralized training, every MAPPO agent observes the complete state $\mathbf{s}$, enabling cooperation among modalities.

Given the shared state, the policy network of agent $c$ generates a continuous action corresponding to the importance of its associated sensing modality,
\(
a_c \sim \pi_{\theta_c}(a_c \mid \mathbf{s}),
\)

where $\pi_{\theta_c}$ denotes the stochastic policy parameterized by $\theta_c$. The actions generated by all agents are normalized using the softmax function to obtain the final fusion weights,
\(
\tilde{w}_c
=
\frac{\exp(a_c)}
{\sum_{j=1}^{C}\exp(a_j)},
\qquad
\sum_{c=1}^{C}\tilde{w}_c=1.
\)

The adaptive multimodal prediction is then obtained by weighting the modality-specific expert logits,
\(\hat{\mathbf{z}}
=
\sum_{c=1}^{C}
\tilde{w}_c
\mathbf{z}_c,
\)
where $\mathbf{z}_c=g_c(\mathbf{e}_c)$ denotes the logit vector produced by the $c$-th expert. Finally, the fused logits are converted into class probabilities using the softmax function for activity prediction.

During training, a centralized critic estimates the state-value function $V_{\phi}(\mathbf{s})$ and evaluates the quality of the joint policy. The critic is optimized using the PPO objective described in Section~3, while the actor networks are updated to maximize the expected cumulative reward based on the quality of the fused prediction.


\paragraph{Reward Function:}

The objective of the MAPPO agents is to learn modality specific fusion weights that maximize classification performance while minimizing the reliance on expensive sensing modalities. To achieve this objective, we design a dense reward function that jointly considers prediction improvement and sensor acquisition cost:

\begin{equation*}
R_t=
\underbrace{\left(\hat{p}^{\mathrm{MAPPO}}_y-\hat{p}^{\mathrm{eq}}_y\right)}_{\text{Accuracy Gain}}
-
\lambda
\underbrace{
\frac{\sum_{c=1}^{C}\tilde{w}_c\kappa_c}
{\sum_{c=1}^{C}\kappa_c}
}_{\text{Normalized Sensor Cost}},
\label{eq:reward}
\end{equation*}

where $\hat{p}^{\mathrm{MAPPO}}_y$ denotes the softmax probability assigned to the ground-truth class $y$ using the adaptive MAPPO-generated fusion weights, while $\hat{p}^{\mathrm{eq}}_y$ is the corresponding probability obtained under the equal-weight fusion strategy introduced in Stage~2. The first term therefore measures the improvement in prediction confidence achieved by adaptive multimodal fusion. A positive value indicates that the learned policy outperforms equal-weight fusion, whereas a negative value penalizes inferior decisions. Since this reward is computed for every training sample, it provides a dense learning signal and alleviates the sparse reward problem commonly encountered in reinforcement learning.

The second term introduces a cost-aware regularization based on the acquisition cost of each sensing modality. Here, $\kappa_c$ denotes the normalized acquisition cost of modality $c$, and $\tilde{w}_c$ is the fusion weight assigned by the MAPPO agent. Consequently, modalities with higher acquisition costs receive a larger penalty when assigned larger fusion weights. The denominator normalizes the total acquisition cost, restricting the cost penalty to the interval $[0,1]$. The trade-off parameter $\lambda\ge0$ controls the balance between recognition accuracy and sensing cost, with larger values encouraging the policy to rely more heavily on lower-cost modalities whenever comparable predictive performance can be achieved.

\begin{algorithm}[t]
\caption{Centralized MAPPO-MoE for Cost-Aware Multimodal HAR}
\label{alg:centralised}
\begin{algorithmic}[1]

\Require Sensor streams $\{\mathcal{D}_c\}_{c=1}^{C}$, sensor costs $\{\kappa_c\}$
\Ensure Cost-aware multimodal fusion policy $\{\pi_c\}$

\Statex \textbf{Stage 1: Independent Expert Pretraining}
\State Train each encoder-expert pair $(f_c,g_c)$ independently on modality $\mathcal{D}_c$

\Statex \textbf{Stage 2: Baseline MoE}
\State Train joint fusion with equal weights:
\[
\hat{\mathbf{y}}
=
\frac{1}{C}
\sum_{c=1}^{C}
g_c(f_c(\mathbf{x}_c))
\]

\Statex \textbf{Stage 3: MAPPO Policy Learning}
\For{each training step}
    \State Compute shared state
    $\mathbf{s} =
    [f_1(\mathbf{x}_1)\Vert\cdots\Vert f_C(\mathbf{x}_C)]$

    \State Each agent $c$ samples fusion weight
    $w_c\sim\pi_c(\mathbf{s})$

    \State Fuse experts:
    $\hat{\mathbf{y}}
    =
    \sum_{c=1}^{C}
    \tilde{w}_c\,g_c(f_c(\mathbf{x}_c))$,
    where
    $\tilde{w}_c=\operatorname{softmax}(\mathbf{w})_c$

    \State Compute reward:
    \[
    r =
    \underbrace{
    \left(
    \hat{p}^{\mathrm{MAPPO}}_y
    -
    \hat{p}^{\mathrm{eq}}_y
    \right)
    }_{\text{accuracy gain}}
    -
    \lambda
    \underbrace{
    \frac{\sum_{c=1}^{C}\tilde{w}_c\kappa_c}
    {\sum_{c=1}^{C}\kappa_c}
    }_{\text{sensor cost}}
    \]

    \State Update $\{\pi_c\}$ and $V$ via PPO
\EndFor

\State \Return $\{\pi_c\}$

\end{algorithmic}
\end{algorithm}

\subsection{Proposed Federated Learning} \label{proposed_fl}
\begin{figure} 
    \centering
    \includegraphics[scale = 0.45]{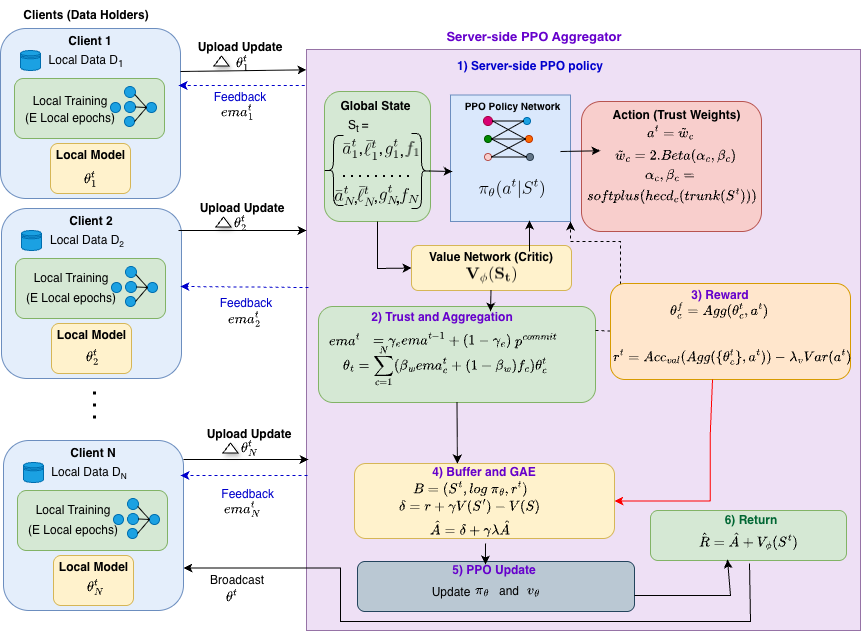}
    \caption{Overview of the BiFL-PPO server-client architecture. Each client reports its local training state $S^t$ (validation accuracy, loss, gradient norm, data-share) to a shared server-side trunk network; per-client heads then produce trust weights $w^t$, which are sent back to each client and simultaneously used to weight aggregation. Transitions $(S^t, \log\pi, r^t)$ are stored in a trajectory buffer and, every three rounds, used to compute Generalized Advantage Estimates and perform a PPO update on the policy and critic.}
    \label{fig:cen_arch_fl}
    \vspace{-3mm}
\end{figure}

The BiFL-PPO, proposed by us for our federated settings, learns a per-client trust score at the server and feeds it back to each client to jointly guide aggregation and local optimization. Fig.~\ref{fig:cen_arch_fl} depicts the overall architecture of this bidirectional client-server interaction.
\begin{algorithm}[t]
\caption{BiFL-PPO: Bidirectional Federated Optimization}
\label{alg:bifl_ppo}
\begin{algorithmic}[1]
\Statex \textbf{Input:} $K$ clients with datasets $\{\mathcal{D}_c\}$; rounds $R$; local epochs $E$; PPO update interval $U{=}3$
\Statex \textbf{Output:} Global model; trust vector $\mathrm{ema}^R$

\State Initialize global model $\theta^0$; policy $\pi_\theta$; critic $V_\phi$
\State Initialize trust $\mathrm{ema}^0 \leftarrow 1/N\ \forall c$; buffer $\mathcal{B} \leftarrow \emptyset$
\For{round $t = 1, \dots, R$}
    \Statex \hspace{1em}\textit{// Local training, using previous round's trust $\mathrm{ema}^{t-1}$}
    \For{each client $c \in \{1,\dots,N\}$ \textbf{in parallel}}
        \State $\theta_c^t \leftarrow \textsc{LocalUpdate}(\theta^{t-1}, \mathcal{D}_c)$, with local learning rate and proximal
        \Statex \hspace{3.4em} strength scaled by $\mathrm{ema}_c^{t-1}$ (higher trust $\rightarrow$ higher LR, weaker penalty)
        \State Client sends $\theta_c^t$ and $a_c^t,\ \ell_c^t,\ g_c^t$ to the server
    \EndFor
    \Statex \hspace{1em}\textit{// Server state}
    \State $\bar a_c^t \leftarrow \beta_s \bar a_c^{t-1}+(1{-}\beta_s)a_c^t$;\quad $\bar\ell_c^t \leftarrow \beta_s \bar\ell_c^{t-1}+(1{-}\beta_s)\ell_c^t\quad \forall c$
    \State $S^t \leftarrow \big[\bar a_c^t,\bar\ell_c^t,g_c^t,f_c\big]_{c=1}^{N}$
    \Statex \hspace{1em}\textit{// Policy and prior blend}
    \State $(\alpha_c,\beta_c) \leftarrow \mathrm{softplus}\big(\mathrm{head}_c(\mathrm{trunk}(S^t))\big)\quad \forall c$
    \State $p^{\text{learned}} \leftarrow \mathrm{softmax}\big(2\cdot\mathbb{E}[\mathrm{Beta}(\alpha_c,\beta_c)]\big)$
    \State $p_c^{\text{prior}} \leftarrow \mathrm{softmax}\big(\beta_p(\bar a_c^t - \max_j \bar a_j^t)\big)$
    \State $p^{\text{commit}} \leftarrow \alpha_p\, p^{\text{prior}} + (1{-}\alpha_p)\, p^{\text{learned}}$
    \State $\mathrm{ema}^t \leftarrow \gamma_e\, \mathrm{ema}^{t-1} + (1{-}\gamma_e)\, p^{\text{commit}}$;\quad $\mathrm{ema}^t \leftarrow \max(\mathrm{ema}^t,\tau_{\min})$
    \Statex \hspace{1em}\textit{// Aggregate (blended with $f_c$) and give bidirectional feedback}
    \State $\theta^t \leftarrow \sum_{c=1}^{N} \big(\beta_w\, \mathrm{ema}_c^t + (1{-}\beta_w)\, f_c\big)\, \theta_c^t$
    \State Report evaluation model as an exponential moving average of $\theta^t$, decay $0.7$ (Training Stability)
    \Statex \hspace{1em}\textit{// Reward: counterfactual aggregation, discarded after use}
    \State Sample $\tilde{w}_c \sim 2\cdot\mathrm{Beta}(\alpha_c,\beta_c)$, forming action $a^t = (\tilde w_1,\dots,\tilde w_N)$; record $\log\pi_\theta(a^t\mid S^t)$
    \State $r^t \leftarrow \mathrm{Acc}_{\text{val}}\big(\mathrm{Agg}(\{\theta_c^t\}, a^t)\big) - \lambda_v\, \mathrm{Var}(a^t)$
    \State $r^t \leftarrow \lambda_r\, r^{t-1} + (1{-}\lambda_r)\, r^t$
    \State Store $(S^t, \log\pi_\theta, r^t)$ in $\mathcal{B}$
    \If{$t \bmod U = 0$}
        \For{each buffered step $\tau$ in $\mathcal{B}$, backward}
            \State $\delta^\tau \leftarrow r^\tau + \gamma V_\phi(S^{\tau+1}) - V_\phi(S^\tau)$
            \State $\hat A^\tau \leftarrow \delta^\tau + \gamma\lambda\, \hat A^{\tau+1}$;\quad $\hat R^\tau \leftarrow \hat A^\tau + V_\phi(S^\tau)$
        \EndFor
        \Statex \hspace{2em} \small ($\gamma{=}0.99,\ \lambda{=}0.95$; $V_\phi$ evaluated on demand from stored $S^\tau$)
        \State Update $\pi_\theta$ via clipped-surrogate PPO loss with entropy bonus, using $\{\hat A^\tau\}$
        \State Update $V_\phi$ via MSE regression onto $\{\hat R^\tau\}$
        \State $\mathcal{B} \leftarrow \emptyset$
    \EndIf
\EndFor
\end{algorithmic}
\end{algorithm}

\subsubsection{Server State Representation}
At the start of round $t$, each client $c \in \{1, \dots, N\}$ sends the server a short summary of how its most recent local training went: its validation accuracy $a_c^t$, its average training loss $\ell_c^t$, and the norm of its gradients $g_c^t$ (clipped and normalized to $[0,1]$ so that no single client can dominate the signal). Together with each client's data-share $f_c = |D_c| / \sum_j |D_j|$, how much of the total training data that client holds, which never changes round to round, these four numbers per client are stacked into a server state vector $S^t \in \mathbb{R}^{4N}$.

A single round's accuracy or loss can be a noisy, unreliable snapshot: a client might have an unusually good or bad batch purely by chance. To avoid the server reacting to this noise, both signals are smoothed using an exponential moving average, each round's value is blended with the running history rather than trusted outright:
\[
\bar{a}_c^t = \beta_s \bar{a}_c^{t-1} + (1-\beta_s) a_c^t, \qquad
\bar{\ell}_c^t = \beta_s \bar{\ell}_c^{t-1} + (1-\beta_s) \ell_c^t,
\]
with smoothing factor $\beta_s = 0.6$, meaning the server's belief about a client updates gradually rather than jumping to match every new observation.

\subsubsection{Formulating Trust Assignment as Reinforcement Learning}
Rather than fixing a hand-designed rule for how much to trust each client, we cast this decision as a standard reinforcement learning problem, with the server acting once per communication round. This requires four ingredients: a \textit{state} describing the current round, an \textit{action} representing a trust decision, a \textit{policy} that chooses actions, and a \textit{reward} telling the policy how good a given decision turned out to be.

\textbf{State.} As described above, the state $S^t \in \mathbb{R}^{4N}$ summarizes every client's recent behaviour, its smoothed accuracy, smoothed loss, gradient norm, and data-share, concatenated across all $N$ clients:
\[
S^t = \big[\bar{a}_1^t, \bar{\ell}_1^t, g_1^t, f_1,\ \dots,\ \bar{a}_N^t, \bar{\ell}_N^t, g_N^t, f_N\big].
\]

\textbf{Action.} The action is a candidate trust weight for every client at once, $a^t = (\tilde{w}_1^t, \dots, \tilde{w}_N^t) \in (0,2)^N$, drawn directly from the policy rather than computed deterministically. This is precisely the stochastic sample $\tilde{w}$ used later to build the counterfactual model for reward computation, action and sampled trust weight are the same quantity, distinct from the smoothed, deterministic trust value that is actually used for aggregation and client feedback.

\textbf{Policy.} The policy defines a probability distribution over actions given the current state, factorized independently across clients:
\[
\pi_\theta(a^t \mid S^t) = \prod_{c=1}^{N} \mathrm{Beta}\!\left(\frac{\tilde{w}_c^t}{2}\,;\ \alpha_c, \beta_c\right), \qquad
\alpha_c, \beta_c = \mathrm{softplus}\big(\mathrm{head}_c(\mathrm{trunk}(S^t))\big).
\]
Sampling an action means drawing each client's weight independently from its own Beta distribution, then scaling by $2$: $\tilde{w}_c^t \sim 2\cdot\mathrm{Beta}(\alpha_c,\beta_c)$.

\textbf{Reward.} The reward measures how well a sampled action would have performed, had it actually been used for aggregation, the resulting model is constructed solely for this purpose and discarded afterward, without affecting the actual aggregated model, penalized to discourage overconfident, degenerate trust assignments:
\[
r^t = \mathrm{Acc}_{\text{val}}\big(\mathrm{Agg}(\{\theta_c^t\}, a^t)\big) - \lambda_v\, \mathrm{Var}(a^t),
\]
further smoothed across rounds via $r^{t}_{\text{smooth}} = \lambda_r\, r^{t-1}_{\text{smooth}} + (1-\lambda_r)\, r^t$ before being used for training.

\textbf{Value function.} A separate critic $V_\phi(S^t)$ estimates the expected discounted return from state $S^t$, used together with the reward to compute the GAE-based advantage that ultimately drives the PPO update described later in this section.

\subsubsection{Policy and Value Networks}
Given this state, the server needs to decide how much to trust each client. Rather than outputting a single trust number directly, the server policy $\pi_\theta$ outputs, for each client, the two parameters of a Beta distribution, essentially a belief about what the right trust value should be, rather than a single committed guess. This shared-trunk, per-client-head design lets the server build one unified understanding of the round while each small head only learns to specialize it into that client's own trust belief, keeping parameters far smaller than $N$ independent networks and naturally supporting cross-client comparisons a per-client-isolated network could not express. All $N$ clients' distributions are produced by a shared network (the ``trunk'') that first builds a general understanding of the whole round, followed by a small client-specific ``head'' that turns that shared understanding into one client's own parameters:
\[
\alpha_c, \beta_c = \mathrm{softplus}\big(\mathrm{head}_c(\mathrm{trunk}(S^t))\big), \qquad
w_c \sim 2 \cdot \mathrm{Beta}(\alpha_c, \beta_c) \in (0, 2).
\]
A Beta distribution is used here instead of the more common Gaussian because it naturally stays within a fixed range on its own, no need to manually clip the output afterward, while still being flexible enough to express either strong, confident trust in a client (a sharply peaked distribution) or genuine uncertainty (a flat, close-to-uniform distribution) depending on what the network has learned. Alongside this policy, the critic network $V_\phi(S^t)$ introduced above learns to estimate how good the current round's situation is overall, which is later used to judge whether a specific trust decision turned out better or worse than expected.

\subsubsection{Trust Computation and Bidirectional Feedback}
The policy's raw output is not used on its own, it is combined with a simple, hand-designed rule of thumb: clients with higher recent accuracy should generally be trusted more. This rule is expressed as a prior probability that favors higher-accuracy clients,
\[
p_c^{\text{prior}} = \frac{\exp(\beta_p (\bar{a}_c^t - \max_j \bar{a}_j^t))}{\sum_k \exp(\beta_p (\bar{a}_k^t - \max_j \bar{a}_j^t))}, \qquad
p^{\text{commit}} = \alpha_p \, p^{\text{prior}} + (1-\alpha_p)\, p^{\text{learned}},
\]
where $p^{\text{learned}} = \mathrm{softmax}\big(2\cdot\mathbb{E}[\mathrm{Beta}(\alpha_c,\beta_c)]\big)$ is the policy's own trust estimate, obtained by normalizing each client's Beta-distribution mean into a probability across clients. The two are blended together, so the learned policy is guided by this common-sense prior rather than starting from nothing. The result is then smoothed across rounds with its own momentum term ($\mathrm{ema}^t = \gamma_e\, \mathrm{ema}^{t-1} + (1-\gamma_e)\, p^{\text{commit}}$) so trust does not swing sharply from one round to the next, and a minimum trust floor $\tau_{\min}$ ensures no client is ever completely excluded, however low its recent performance.

This final trust value is used in two places at once, which is what makes the scheme \textit{bidirectional}. First, it determines the server-side aggregation weight for each client, blended with that client's static data-share so that trust does not entirely override the amount of data a client actually contributes:
\[
\theta^t = \sum_{c=1}^{N} \big(\beta_w\, \mathrm{ema}_c^t + (1-\beta_w)\, f_c\big)\, \theta_c^t,
\]
which is the aggregation rule referenced as $\mathrm{Agg}(\cdot)$ in the reward computation above. Second, this same trust value is fed back to each client to shape its own next round of local training, a more trusted client gets a higher effective learning rate, while a less trusted one is pulled more strongly back toward the current global model through a stronger proximal penalty. In effect, a client the server currently doubts is both counted for less at aggregation time and held on a tighter leash during its own training.
\subsubsection{PPO Update}
Rather than updating after every communication round, the server accumulates experience over three rounds before performing a PPO update. For each buffered transition $t$, the temporal-difference residual is
\[
\delta^t = r^t + \gamma\, V_\phi(S^{t+1}) - V_\phi(S^t),
\]
and Generalized Advantage Estimation (GAE) combines these residuals with exponential decay $\gamma\lambda$ across the buffered window:
\[
\hat{A}^t = \delta^t + \gamma\lambda\, \hat{A}^{t+1}, \qquad \hat{R}^t = \hat{A}^t + V_\phi(S^t),
\]
computed recursively backward through the buffer ($\gamma=0.99,\ \lambda=0.95$), where $\hat{R}^t$ is the return target used to train the critic. The policy is then updated using PPO's clipped-surrogate objective with an entropy bonus to encourage exploration, while the critic is trained via MSE regression toward $\hat{R}^t$.

\subsubsection{Training Stability}
To improve training stability, three additional measures are adopted. First, optimizer momentum is preserved across communication rounds to maintain learning dynamics. Second, each client's learning rate decays gradually to $30\%$ of its initial value by the final round. Third, model performance is reported using an exponential moving average (Polyak averaging, decay $0.7$) of the global model, providing a more stable estimate than the raw aggregated model.\\

We apply our proposed approach to every federated aggregation method in Sec.~\ref{fed_aggr}. Local client optimization remains unchanged; only the aggregation weight is modified, replacing $p_k = n_k/n$ with $\alpha_k = \beta\,\hat{w}_k + (1-\beta)\,p_k$, a blend of $p_k$ and a trust weight $\hat{w}_k$ learned by a server-side PPO agent from each client's local accuracy, loss, and gradient norm ($\beta = 0.85$), trained on a reward from the counterfactual validation accuracy of a sampled weight assignment, penalized by its variance and smoothed across rounds. Substituting $\alpha_k$ for $p_k$ in each method's aggregation rule, the weighted sum for FedAvg, FedProx, and AdaFedProx; $\sum_k \alpha_k d_k$ with $\tau_{\text{eff}} = \sum_k \alpha_k \tau_k$ for FedNova; and the non-BN average for FedBN, gives a uniform PPO-guided aggregation across all five baselines while preserving each method's own local objective and aggregation structure.

\section{Experimental Setup}
All experiments were conducted in Python 3.12 using PyTorch 2.12 on a single NVIDIA Tesla T4 GPU (16 GB VRAM). The proposed centralized MAPPO-MoE framework and federated FL PPO FedAvg model were implemented from scratch, with all reinforcement learning components developed using native PyTorch modules. For federated learning, a sensor-heterogeneous client partition strategy was adopted, where each sensor modality was treated as an independent client and only the expert classifier heads were aggregated globally. Baseline methods including \textit{FedAvg}, \textit{FedProx}, \textit{FedBN}, \textit{AdaFedProx}, \textit{FedNova}, and \textit{FedDPO} were implemented under the same architecture and evaluation protocol to ensure a fair comparison.

\begin{table}[ht]
\centering
\caption{Hyperparameter Settings for Centralized and Federated Learning}
\label{tab:hyperparameters}

\scriptsize

\begin{tabular}{|l|c|c|}
\hline
\textbf{Hyperparameter} & \textbf{Centralized} & \textbf{Federated} \\
\hline
Pretraining Epochs & 40 & -- \\
\hline
MoE Training Epochs & 40 & -- \\
\hline
MAPPO Training Epochs & 60 & -- \\
\hline
Communication Rounds & -- & 50 \\
\hline
Local Epochs (MEx) & -- & 3 \\
\hline
Local Epochs (UTD-MHAD) & -- & 2 \\
\hline
Model Learning Rate & $5\times10^{-4}$ & $5\times10^{-4}$ \\
\hline
PPO Actor Learning Rate & $1\times10^{-4}$ & $3\times10^{-4}$ \\
\hline
PPO Critic Learning Rate & $1\times10^{-4}$ & $3\times10^{-4}$ \\
\hline
PPO Clip ($\epsilon$) & 0.2 & 0.2 \\
\hline
Entropy Coefficient & 0.01 & 0.05 \\
\hline
Discount Factor ($\gamma$) & 0.99 & 0.99 \\
\hline
GAE Parameter ($\lambda$) & 0.95 & 0.95 \\
\hline
Cost Weight ($\lambda_{cost}$) & 0.3 & 0.3 \\
\hline
Rollout Buffer Size & 32 & 16 \\
\hline
Gradient Clipping & 1.0 & 1.0 \\
\hline
Label Smoothing & 0.1 & 0.1 \\
\hline
\end{tabular}

\vspace{-3mm}
\end{table}

\subsection{Dataset Description}
This subsection presents a brief description of the two datasets used in \textit{FedMHAR}, namely MEx and UTD MHAD.
\subsubsection{MEx Dataset}
The MEx (Multi-modal Exercise) dataset \cite{wijekoon2019MEx} is a publicly available Human Activity Recognition (HAR) dataset designed for exercise recognition and rehabilitation monitoring. It contains data from 30 subjects performing 7 physiotherapy exercises using multiple sensor modalities, including wearable accelerometers, a pressure mat, and a depth camera. The dataset is widely used for research in multimodal sensor fusion, activity recognition, and deep learning-based HAR systems.

\subsubsection{\textit{UTD MHAD} Dataset}
The UTD-MHAD (University of Texas at Dallas Multimodal Human Action Dataset)~\cite{7350781} is a benchmark dataset for multimodal human activity recognition, comprising 27 actions performed by 8 subjects with synchronized RGB, depth, skeleton, and inertial data. In this work, we use the 3D skeletal joint positions and inertial signals (accelerometer and gyroscope), while excluding RGB and depth modalities to maintain a lightweight and privacy-preserving framework.

In addition, we derive a third modality, joint velocity, by normalizing the skeleton sequence and then taking its frame-to-frame difference:
\[
\hat{p}_j^{(t)} = \frac{p_j^{(t)} - \mu_j}{\sigma_j}, \qquad
v_{j}^{(t)} =
\begin{cases}
0, & t = 1 \\[2pt]
\hat{p}_{j}^{(t)} - \hat{p}_{j}^{(t-1)}, & t = 2, \dots, T,
\end{cases}
\]
where $p_j^{(t)} \in \mathbb{R}^3$ denotes the 3D position of joint $j$ at frame $t$, and $\hat{p}_j^{(t)}$ its z-score normalized coordinate over the sequence. The first frame's velocity is set to zero rather than shortening the sequence, keeping joint velocity the same length as the other modalities. Joint velocity captures motion dynamics complementary to static joint positions, without requiring additional sensors. Thus, our framework utilizes three modalities: skeleton position, joint velocity, and inertial signals.

\section{Results \& Discussion}
This section presents the results of \textit{FedMHAR} in both centralized and federated settings, organized by the two datasets, and compares its performance with state-of-the-art federated learning methods.

\subsection{Centralized Setup}
For the centralized setup, the different modalities of the multimodal datasets are fused using the Mixture of Experts (MoE) framework, while MAPPO performs weighted fusion to determine the importance of each modality for recognizing a specific activity.

\subsubsection{MEx Dataset}

\begin{figure*}[ht!]
\centering

\subfigure[]{
\includegraphics[width=0.47\linewidth]{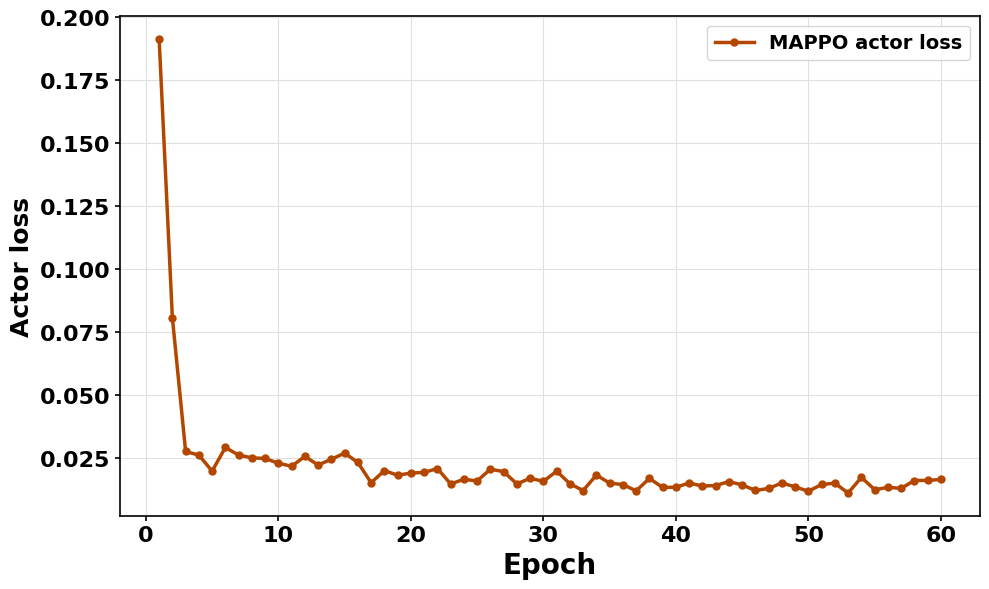}
}
\hfill
\subfigure[\label{kl_mqtt_len}]{
\includegraphics[width=0.47\linewidth]{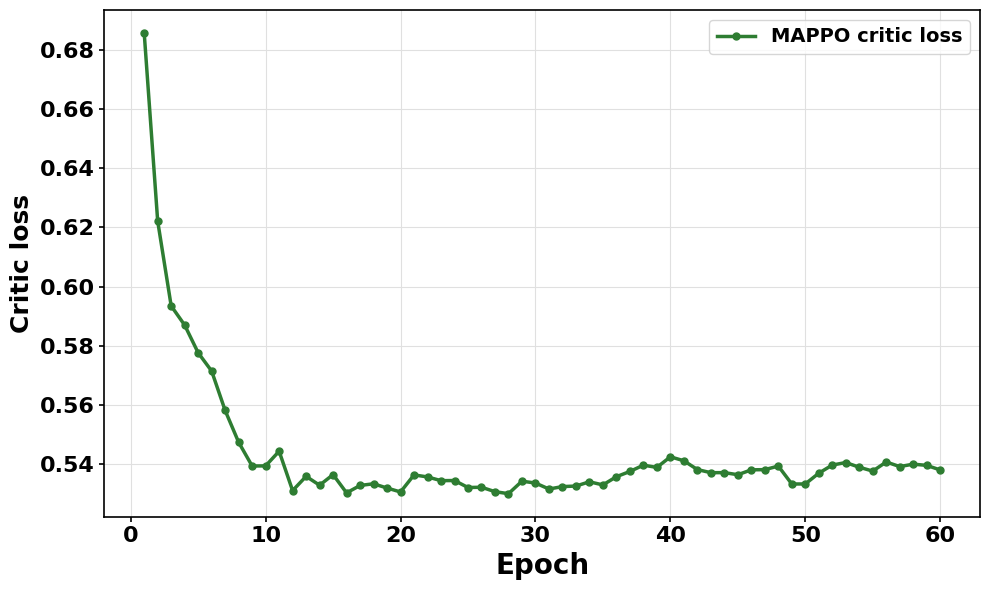}
}

\caption{(a) Loss value of the actor network of MAPPO for MEx dataset; (b) Loss value of the critic network of MAPPO for MEx dataset;}
\label{fig:loss_MEx}

\vspace{-3mm}
\end{figure*}

Fig.~\ref{fig:loss_MEx}(a) and Figure~\ref{fig:loss_MEx}(b) show the MAPPO actor and critic losses on the MEx dataset after applying return normalization and a cosine-annealed critic learning rate. The actor loss decreases rapidly from approximately $0.191$ to $0.017$ by epoch 60, while the critic loss drops from approximately $0.685$ to a stable plateau around $0.53$. Both losses converge within the first approx. 10 epochs and remain stable thereafter, demonstrating faster and smoother optimization than the earlier configuration. The improved stability is attributed to return normalization and the adaptive critic learning rate, which provide better-conditioned value targets and lower-variance advantage estimates, resulting in more stable policy updates.

\begin{figure} 
    \centering
    \includegraphics[scale = 0.3]{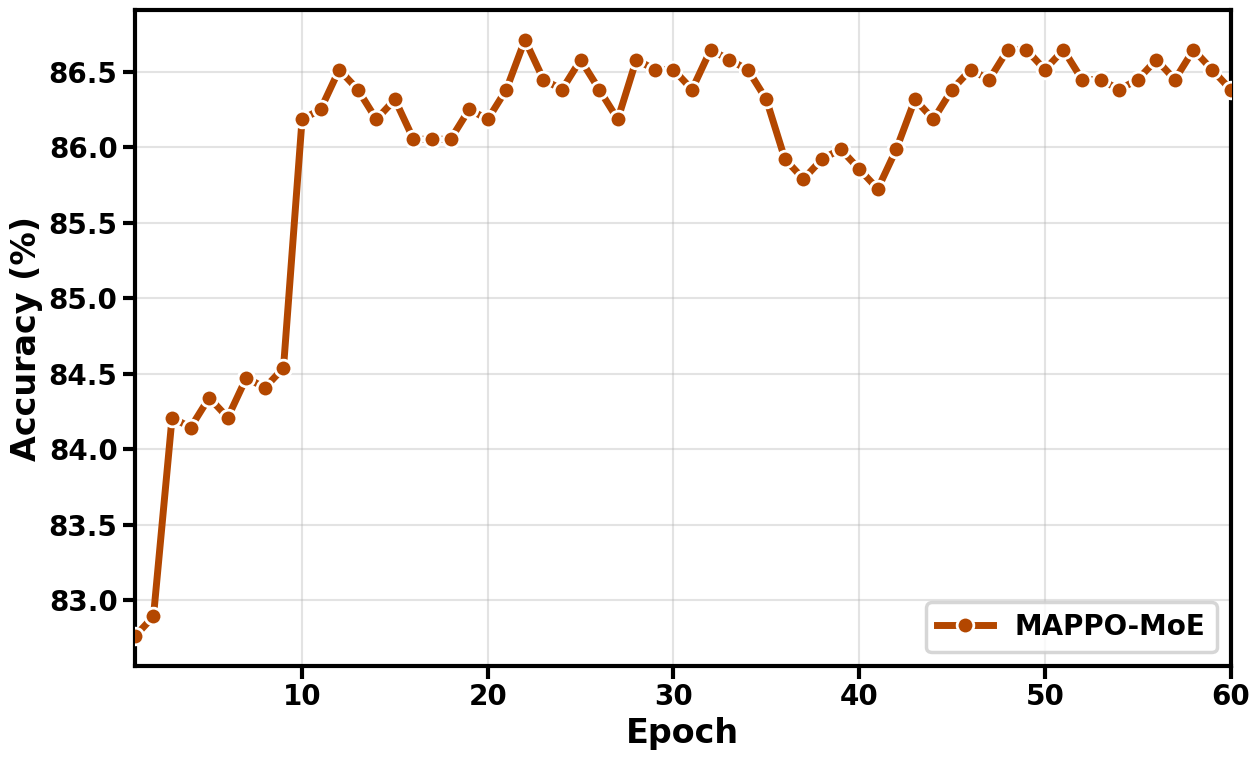}
    \caption{The Improvement of the MAPPO-Based Centralized Architecture with respect to the Baseline MoE for MEx Dataset}
    \label{fig:mappo_MEx}
    \vspace{-3mm}
\end{figure}

Fig. \ref{fig:mappo_MEx} compares the test accuracy of the proposed MAPPO-MoE with the Baseline MoE on the MEx dataset. The Baseline MoE achieves an accuracy of 82.24\% using equal-weight fusion of all sensor modalities, whereas MAPPO-MoE consistently maintains higher accuracy throughout training and attains a peak accuracy of 89.47\%, corresponding to an improvement of 7.24 percentage points. The performance gain is attributed to the adaptive modality fusion mechanism of MAPPO-MoE, where PPO agents dynamically learn the importance of individual sensor modalities based on their contribution to activity recognition. Unlike static equal-weight fusion, the proposed approach emphasizes the most informative sensors for each exercise while reducing the influence of less relevant modalities, resulting in more discriminative feature representations and improved classification performance across the physiotherapy activities in the MEx dataset.

\begin{table*}[h]
\centering
\caption{Classification Report for MAPPO MoE}
\begin{tabular}{lccc}
\hline
\textbf{Exercise} & \textbf{Precision} & \textbf{Recall} & \textbf{F1-score} \\
\hline
Knee-rolling    & 0.81 & 0.91 & 0.86 \\
Bridging        & 0.83 & 0.80 & 0.82 \\
Pelvic-tilt     & 0.93 & 0.83 & 0.88 \\
The-clam        & 0.88 & 0.88 & 0.88 \\
Ext-in-lying    & 0.98 & 0.94 & 0.96 \\
Prone-punches   & 0.93 & 0.81 & 0.87 \\
Superman        & 0.84 & 1.00 & 0.91 \\
\hline
\label{tab:pre_f1_rec_MEx}
\end{tabular}
\end{table*}

Table \ref{tab:pre_f1_rec_MEx} presents the per-class precision, recall, and F1-score of MAPPO-MoE on the MEx dataset, providing a detailed assessment of its performance across all seven physiotherapy exercises.

The MAPPO-MoE achieves strong and balanced performance across all exercise classes, with the highest F1-score obtained for Ext-in-Lying (0.96). Most classes achieve F1-scores above 0.85, demonstrating the model's ability to effectively distinguish between physiotherapy exercises with subtle biomechanical differences. The comparatively lower performance on Bridging (0.82) reflects the inherent similarity between certain exercises and the variability in their execution. Overall, the results show that adaptive sensor fusion enables the model to exploit the most informative sensor placements for each exercise, leading to a 7.24\% improvement over the equal-weight Baseline MoE and highlighting the effectiveness of RL-driven modality weighting for physiotherapy activity recognition.

\begin{table}[ht]
\centering
\caption{Comparison of Baseline MoE and MAPPO MoE Fusion on the \textit{MEx} Dataset (5-Seed Evaluation)}
\label{tab:MEx_centralized_5seed}
\setlength{\tabcolsep}{10pt}
\renewcommand{\arraystretch}{1.3}
\begin{tabular}{l c}
\toprule
\textbf{Method} & \textbf{Test Accuracy (\%, Mean $\pm$ Std)} \\
\midrule
Baseline MoE & 85.07 $\pm$ 4.04 \\
MAPPO MoE (Ours) & \textbf{87.30 $\pm$ 2.53} \\
\bottomrule
\end{tabular}
\end{table}

We evaluate the centralized fusion model across five random seeds (42, 7, 123, 99, 2024), with results summarized in Table~\ref{tab:MEx_centralized_5seed}. The proposed MAPPO-based MoE fusion achieves a mean test accuracy of 0.87 $\pm$ 0.03, outperforming the equally weighted Baseline MoE (0.85 $\pm$ 0.04) by 0.023. In addition to improving accuracy, MAPPO reduces performance variance, lowering the standard deviation from 0.04 to 0.03. The MAPPO model outperforms the baseline in four of the five seeds, demonstrating that adaptive policy-learned fusion yields both higher accuracy and more stable performance than uniform expert aggregation.

\subsubsection{UTD MHAD Dataset}

\begin{figure*}[ht!]
\centering

\subfigure[]{
\includegraphics[width=0.47\linewidth]{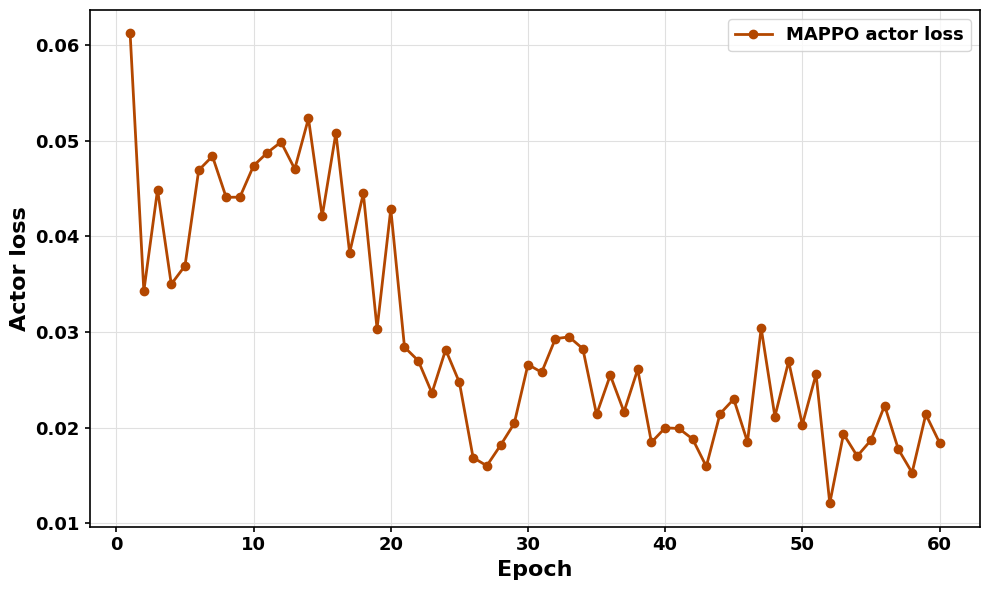}
}
\hfill
\subfigure[\label{kl_mqtt_len}]{
\includegraphics[width=0.47\linewidth]{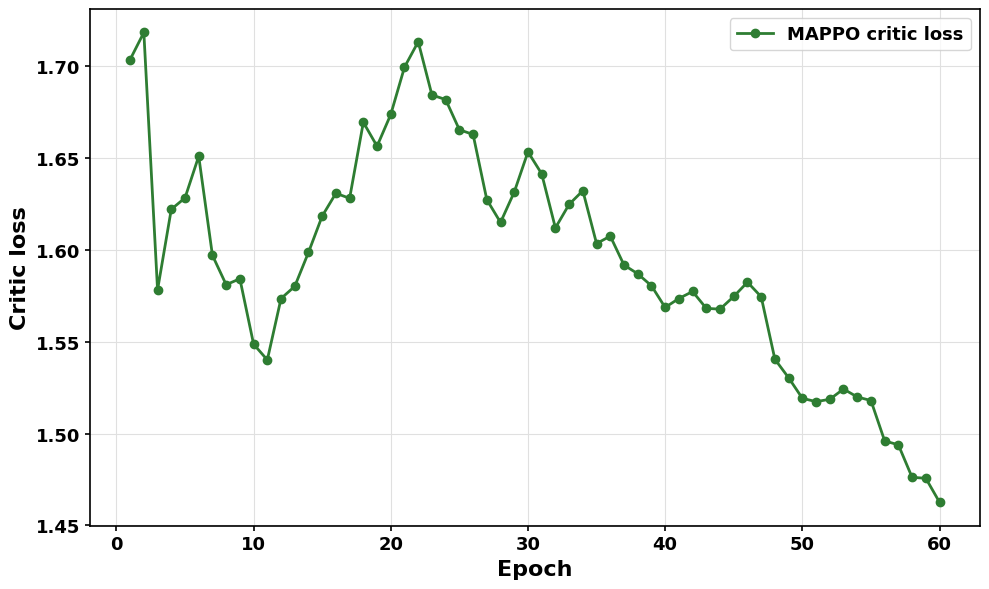}
}

\caption{(a) Loss value of the actor network of MAPPO for UTD MHAD dataset; (b) Loss value of the critic network of MAPPO for UTD MHAD dataset;}
\label{fig:loss_utd}

\vspace{-3mm}
\end{figure*}

Fig.~\ref{fig:loss_utd} (a) and Fig.~\ref{fig:loss_utd} (b) present the MAPPO actor and critic losses averaged over five seeds across 60 training epochs. The actor loss decreases from $0.0612$ to $0.0184$, while the critic loss decreases from $1.7037$ to $1.4626$. Both losses exhibit higher fluctuations during the first 20 epochs before converging to a stable downward trend, reflecting the coupled learning dynamics of the actor and critic as value estimates become more reliable. The remaining fluctuations arise from PPO's minibatch updates and the stochastic reward function combining classification accuracy and sensor-cost terms. Since these losses are reward-based optimization objectives rather than classification losses, they should be interpreted relative to their convergence behavior rather than their absolute magnitudes.

\begin{figure} 
    \centering
    \includegraphics[scale = 0.4]{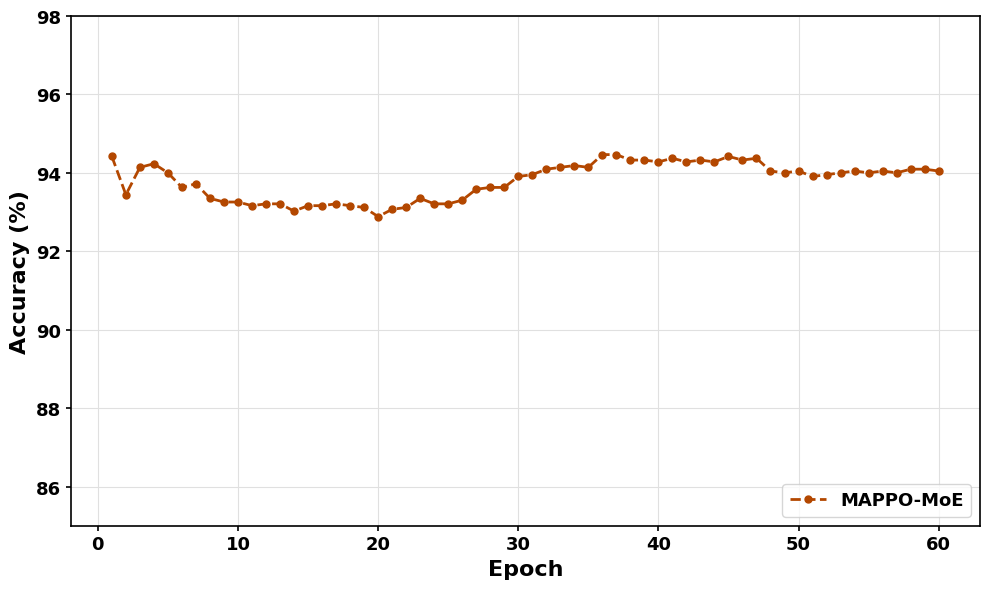}
    \caption{The Improvement of the MAPPO-Based Centralized Architecture with respect to the Baseline MoE for UTD MHAD Dataset}
    \label{fig:mappo_utd}
    \vspace{-3mm}
\end{figure}

Fig. \ref{fig:mappo_utd} represents the performance of the MAPPO-based MoE with respect to the baseline MoE for the \textit{\textit{UTD MHAD}} dataset. The MAPPO-MoE achieves a $5.12\%$ improvement over the Baseline MoE on UTD-MHAD due to its ability to adaptively weight sensor modalities according to their relevance for different action types. UTD-MHAD contains diverse activities ranging from fine hand gestures to full-body movements, where the discriminative power of skeleton, inertial, and joint velocity modalities varies considerably. While the Baseline MoE assigns equal importance to all modalities, MAPPO learns dynamic modality weights through reinforcement learning, emphasizing inertial signals for hand-centric actions, skeleton information for posture-based activities, and joint velocity features for dynamic movements.


\begin{table*}[ht]
\centering
\caption{Classification Report for MAPPO MoE}
\label{tab:pre_f1_rec_utd}
\small
\begin{tabular}{lccc|lccc}
\hline
\textbf{Class} & \textbf{P} & \textbf{R} & \textbf{F1} &
\textbf{Class} & \textbf{P} & \textbf{R} & \textbf{F1} \\
\hline

swipe-left       & 0.89 & 1.00 & 0.94 &
arm-curl         & 1.00 & 1.00 & 1.00 \\

swipe-right      & 0.89 & 1.00 & 0.94 &
tennis-serve     & 1.00 & 1.00 & 1.00 \\

wave             & 0.84 & 1.00 & 0.91 &
push             & 0.94 & 1.00 & 0.97 \\

clap             & 0.94 & 1.00 & 0.97 &
knock            & 0.92 & 0.75 & 0.83 \\

throw            & 1.00 & 0.88 & 0.93 &
catch            & 1.00 & 0.62 & 0.77 \\

arm-cross        & 1.00 & 1.00 & 1.00 &
pickup-throw     & 1.00 & 0.88 & 0.93 \\

basketball-shoot & 1.00 & 1.00 & 1.00 &
jog              & 1.00 & 1.00 & 1.00 \\

draw-x           & 0.89 & 1.00 & 0.94 &
walk             & 1.00 & 1.00 & 1.00 \\

draw-circle-cw   & 1.00 & 1.00 & 1.00 &
sit2stand        & 1.00 & 1.00 & 1.00 \\

draw-circle-ccw  & 1.00 & 1.00 & 1.00 &
stand2sit        & 1.00 & 1.00 & 1.00 \\

draw-triangle    & 1.00 & 1.00 & 1.00 &
forward-lunge    & 0.89 & 1.00 & 0.94 \\

bowling          & 0.84 & 1.00 & 0.91 &
squat            & 1.00 & 1.00 & 1.00 \\

boxing           & 1.00 & 1.00 & 1.00 &
& & & \\

baseball-swing   & 1.00 & 1.00 & 1.00 &
& & & \\

tennis-swing     & 1.00 & 0.81 & 0.90 &
& & & \\

\hline
\end{tabular}
\end{table*}




Table \ref{tab:pre_f1_rec_utd} presents the per-class precision, recall, and F1-score of the proposed MAPPO-MoE model on the UTD-MHAD dataset, providing a detailed evaluation of its classification performance across all 27 activity classes. The MAPPO-MoE demonstrates strong and consistent performance on UTD-MHAD, achieving perfect F1-scores for 14 of the 27 action classes and near-perfect performance for most of the remaining classes. The model effectively recognizes both fine-grained hand gestures and full-body activities by adaptively leveraging complementary information from skeleton, inertial, and joint velocity modalities. The few remaining errors are concentrated in inherently challenging actions such as knock and catch, which exhibit high inter-class similarity and intra-class variability. Overall, the high weighted F1-score of 0.96 highlights the effectiveness of the proposed cost-aware multi-agent fusion strategy in learning robust representations across diverse human activities.

\begin{table}[ht]
\centering
\caption{Comparison of Baseline MoE and MAPPO MoE Fusion on the \textit{UTD-MHAD} Dataset (5-Seed Evaluation)}
\label{tab:utd_centralized_5seed}
\setlength{\tabcolsep}{10pt}
\renewcommand{\arraystretch}{1.3}
\begin{tabular}{l c}
\toprule
\textbf{Method} & \textbf{Test Accuracy (\%, Mean $\pm$ Std)} \\
\midrule
Baseline MoE & 82.84 $\pm$ 0.67 \\
\textbf{MAPPO MoE (Ours)} & \textbf{94.98 $\pm$ 0.88} \\
\midrule
\textbf{Mean Improvement} & \textbf{+12.14} \\
\bottomrule
\end{tabular}
\end{table}

We evaluate the centralized multimodal fusion framework across five random seeds (42, 123, 2024, 7, 99) on the UTD-MHAD dataset, with results summarised in Table~\ref{tab:utd_centralized_5seed}. The proposed MAPPO-based mixture-of-experts fusion achieves a mean test accuracy of $0.95 \pm 0.01$, outperforming the uniformly-weighted Baseline MoE $(0.83 \pm 0.01)$ by a substantial margin of $+0.12$ averaged across all seeds. This gain is consistent and large, MAPPO improves over the Baseline by between $+0.11$ and $+0.14$ depending on the seed, with no seed where the Baseline comes close to the MAPPO result. The low standard deviations of both methods ($0.01$ for Baseline and $0.01$ for MAPPO) confirm that these results are highly reproducible and not dependent on fortunate initialization.
\subsection{Federated Learning Setup}

\subsubsection{MEx Dataset}

\begin{figure*}[ht!]
\centering

\subfigure[]{
\includegraphics[width=0.47\linewidth]{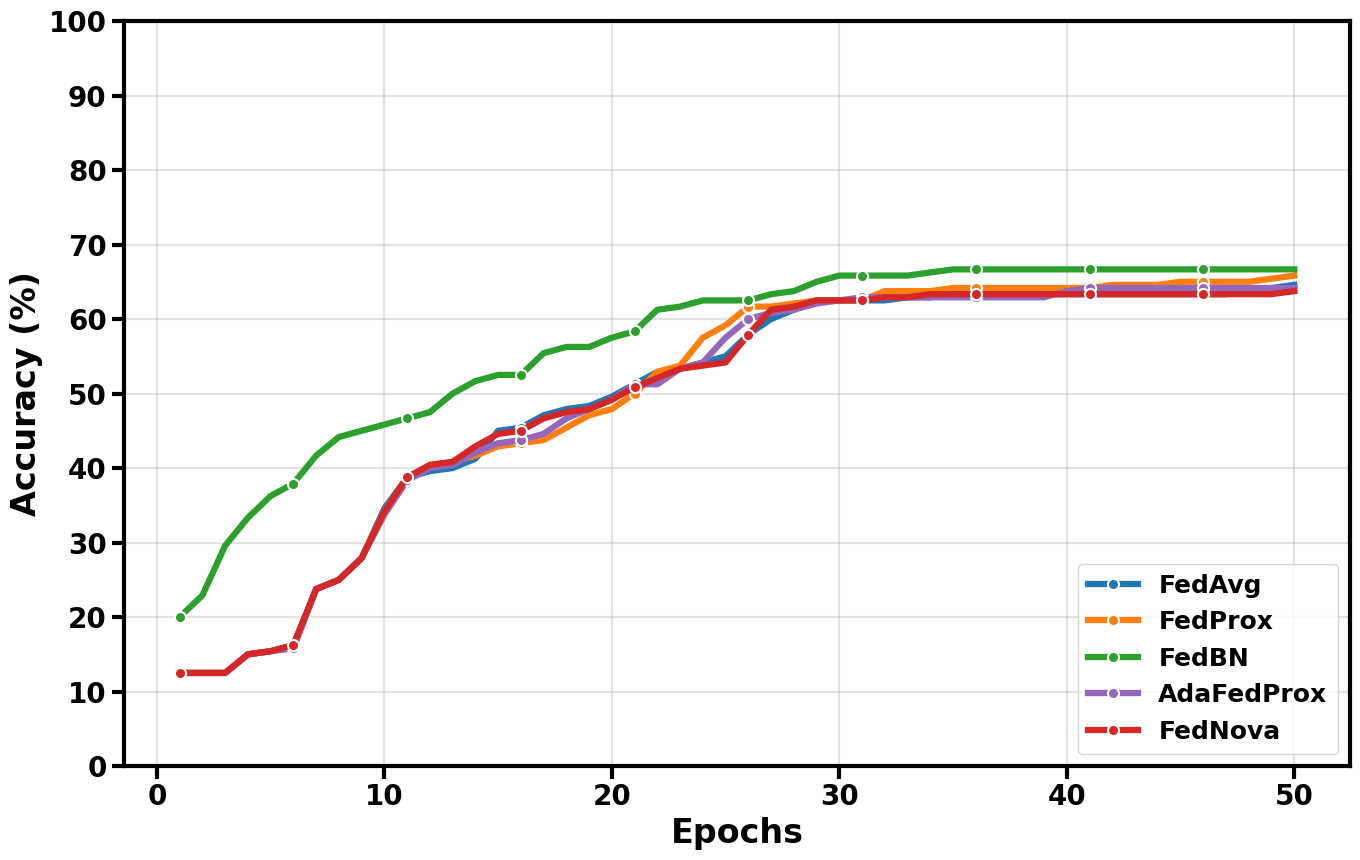}
}
\hfill
\subfigure[\label{kl_mqtt_len}]{
\includegraphics[width=0.47\linewidth]{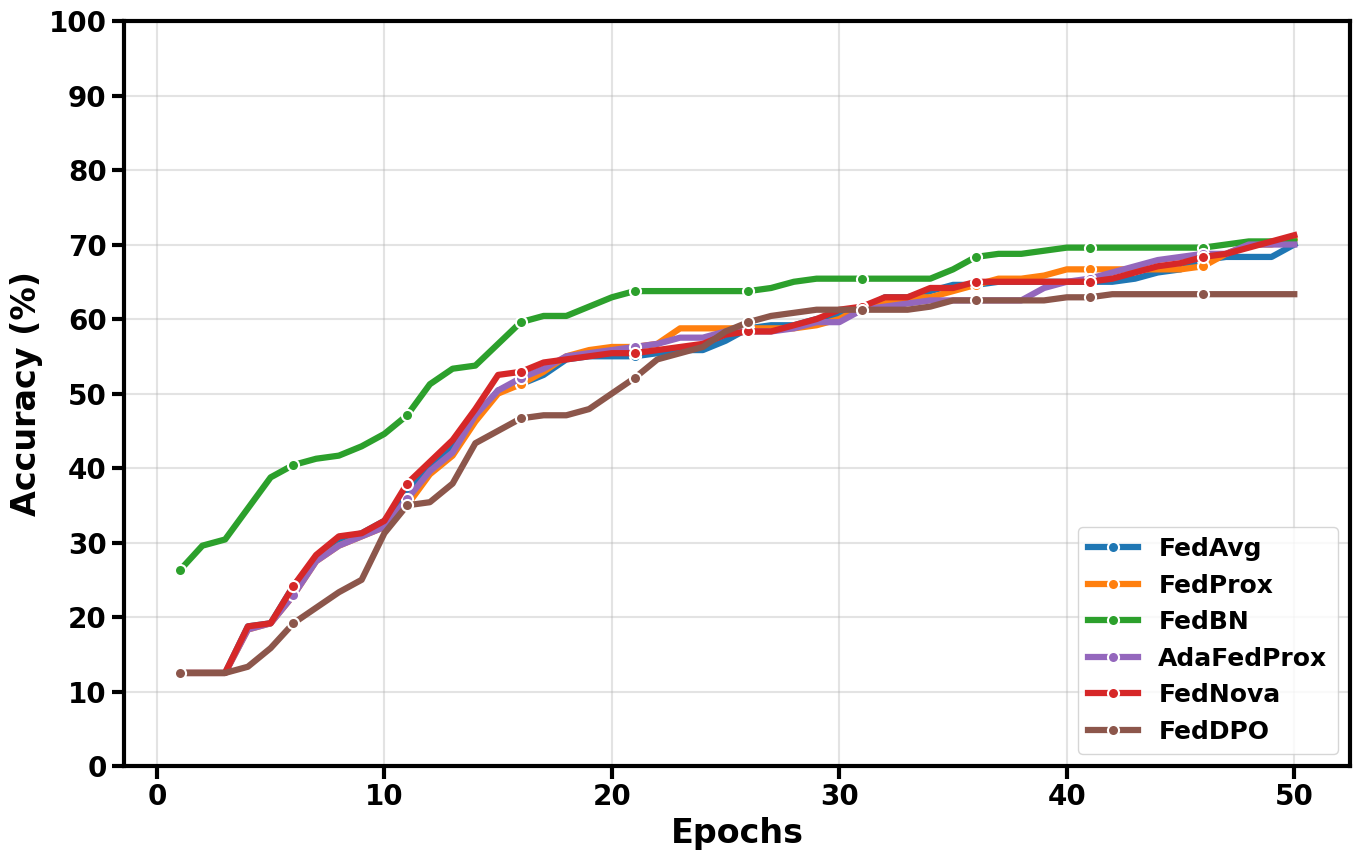}
}

\caption{(a) Test Accuracy of SOTA FL methods over epochs (b) Test Accuracy of SOTA FL with PPO over epochs.}
\label{fig:fl_sota_ppo}

\vspace{-3mm}
\end{figure*}

Fig.~\ref{fig:fl_sota_ppo} (a) and Fig.~\ref{fig:fl_sota_ppo} (b) show epoch-wise test accuracy over 50 communication rounds for the vanilla and PPO-enhanced variants of the five baseline aggregation methods, respectively. In the vanilla setting, all five methods converge to a narrow band between $63\%$ and $67\%$ by round 50, with FedBN and FedProx reaching the highest final accuracy (approx. 66--67\%) and FedAvg, AdaFedProx, and FedNova converging slightly lower (approx. 63--64\%). With our proposed PPO-based aggregation, every method improves over its vanilla counterpart: FedNova shows the largest gain ($+8\%$, from $64\%$ to $72\%$), followed by FedAvg ($+5\%$), FedBN and AdaFedProx ($+4\%$ each), and FedProx showing the smallest but still positive improvement ($+1\%$). This consistent, positive gain across all five methods, despite their differing local optimization strategies (proximal regularization, step normalization, local batch-norm), supports our central claim that adaptively re-weighting client contributions via a learned, server-side trust policy provides a benefit that is complementary to, rather than a substitute for, each method's existing heterogeneity-handling mechanism.

We additionally observe that FedBN converges fastest in early rounds in both figures (reaching approx. 50\% accuracy by round 11, versus 39\% for the other four methods), consistent with its client-local Batch Normalization layers adapting quickly to feature-distribution heterogeneity across clients, a correction the other methods' gradient-based mechanisms achieve only more gradually. FedDPO (Fig.~\ref{fig:fl_sota_ppo} (b)), which uses a separate reinforcement-learning policy to directly decide how strongly each client should be pulled back toward the global model during its own local training, rather than adaptively re-weighting client contributions at server-side aggregation, plateaus near $63$--$64\%$ from round 30 onward, below all five PPO-aggregated methods, suggesting that server-side, trust-weighted aggregation is a more effective use of a learned policy in this setting than client-side proximal tuning alone.

\begin{table}[ht]
\centering
\caption{Bidirectional Server-PPO (BiFL-PPO) vs.\ FedAvg on \textit{MEx} Dataset
         (5-seed, non-IID Dirichlet $\alpha{=}0.5$).}
\label{tab:MEx_bifl_5seed}
\setlength{\tabcolsep}{8pt}
\renewcommand{\arraystretch}{1.3}
\begin{tabular}{l c c}
\toprule
\textbf{Method}
& \textbf{Mean}
& \textbf{Std} \\
\midrule
FedAvg
& 78.09
& 4.83 \\
\textbf{BiFL-PPO (Ours)}
& \textbf{79.74}
& \textbf{0.98} \\
\midrule
\textit{Mean improvement (BiFL-PPO $-$ FedAvg)}
& \textbf{+1.64}
& -- \\
\bottomrule
\end{tabular}
\end{table}

 We evaluate the proposed Bidirectional Federated Learning framework with Server-PPO aggregation (BiFL-PPO) against the FedAvg baseline under a non-IID Dirichlet label split ($\alpha{=}0.5$) across five random seeds, with results summarised in Table~\ref{tab:MEx_bifl_5seed}. BiFL-PPO achieves a mean test accuracy of $79.74\%$ compared to $78.09\%$ for FedAvg, a mean gain of $+1.64$ percentage points. More notably, the standard deviation of BiFL-PPO across seeds is $0.98\%$, nearly five times smaller than FedAvg's $4.83\%$, demonstrating substantially more consistent convergence under heterogeneous data conditions. BiFL-PPO outperforms FedAvg on four of the five seeds; the single exception (seed 2024, $-4.28\%$) corresponds to a Dirichlet partition that assigns one client an unusually representative shard, an outlier condition that gives FedAvg's uniform averaging an atypical advantage. The stability gain is the primary practical contribution of the bidirectional design: by feeding the PPO-learned per-client trust weights back to clients as adaptive learning-rate and proximal-penalty signals, BiFL-PPO suppresses the client drift that causes FedAvg to produce high-variance results across seeds, yielding a more reliable global model under the non-IID data heterogeneity that characterises real-world federated deployments.

\begin{figure} 
    \centering
    \includegraphics[scale = 0.35]{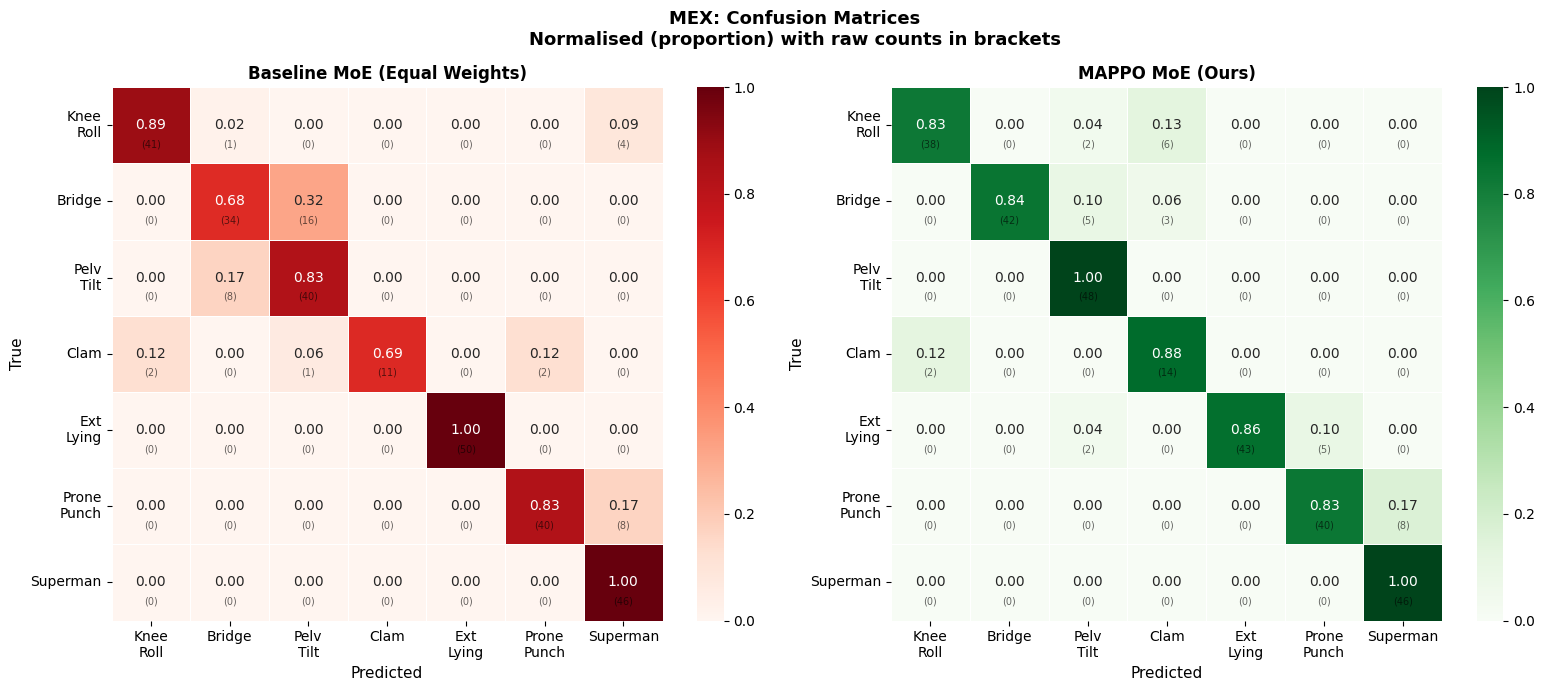}
    \caption{Confusion matrices of Baseline MoE and MAPPO-MoE on the MEx dataset}
    \label{fig:con_MEx}
    \vspace{-3mm}
\end{figure}

Fig. \ref{fig:con_MEx} represents the normalized confusion matrices for the Baseline MoE and MAPPO-MoE on the MEx dataset. The confusion matrices highlight the class pairs that are most difficult to distinguish and demonstrate the effectiveness of the proposed MAPPO-based fusion strategy. The baseline MoE exhibits substantial confusion between Bridge and Pelvic Tilt (32\% and 17\%, respectively) and relatively low recognition of Clam (69\%). In contrast, MAPPO-MoE significantly improves the true positive rates of these challenging classes, increasing Bridge recognition from 68\% to 84\%, Clam from 69\% to 88\%, and Pelvic Tilt from 83\% to 100\%. Although minor degradations are observed for Knee Roll and Extension in Lying, the overall confusion is reduced, indicating that MAPPO learns more discriminative modality fusion and better distinguishes exercises with similar motion patterns.

\begin{table*}[h]
\centering
\caption{Confusion Matrix and Per-Class Metrics for MAPPO MoE (MEx)}
\label{tab:confusion_matrix_MEx}
\resizebox{\textwidth}{!}{%
\begin{tabular}{lcccccccccc}
\hline
\textbf{True $\backslash$ Pred} & \textbf{Knee-roll.} & \textbf{Bridging} & \textbf{Pelvic-ti.} & \textbf{The-clam} & \textbf{Ext-in-ly.} & \textbf{Prone-pun.} & \textbf{Superman} & \textbf{Precision} & \textbf{Recall} & \textbf{F1-score} \\
\hline
\textbf{Knee-roll.} & 170 & 0 & 1 & 5 & 0 & 0 & 0 & 1.00 & 0.97 & 0.98 \\
\textbf{Bridging} & 0 & 160 & 3 & 12 & 0 & 2 & 2 & 0.79 & 0.89 & 0.84 \\
\textbf{Pelvic-ti.} & 0 & 43 & 78 & 33 & 5 & 3 & 11 & 0.93 & 0.45 & 0.61 \\
\textbf{The-clam} & 0 & 0 & 2 & 158 & 0 & 0 & 0 & 0.76 & 0.99 & 0.86 \\
\textbf{Ext-in-ly.} & 0 & 0 & 0 & 1 & 182 & 0 & 0 & 0.96 & 0.99 & 0.98 \\
\textbf{Prone-pun.} & 0 & 0 & 0 & 0 & 2 & 143 & 33 & 0.97 & 0.80 & 0.88 \\
\textbf{Superman} & 0 & 0 & 0 & 0 & 0 & 0 & 173 & 0.79 & 1.00 & 0.88 \\
\hline
\end{tabular}%
}
\end{table*}

Table \ref{tab:confusion_matrix_MEx} reports the confusion matrix and per class metrics for MAPPO MoE on MEx across seven exercises. Of the 1222 test samples, 1064 are correctly classified, giving an overall accuracy of 87.07 percent. Pelvic tilt is the primary source of error, with only 45 percent recall, mostly confused with Bridging, 43 samples, and The clam, 33 samples. This also lowers precision for Bridging and Superman to 79 percent each, since many Pelvic tilt and Prone punches samples are misclassified into these two classes. The remaining exercises, Knee rolling, The clam, Extension in lying, and Superman, all exceed 96 percent recall, indicating that Pelvic tilt is the main source of confusion while the rest of the classes are classified reliably.

\begin{sidewaystable*}[h]
\centering
\caption{Confusion Matrix and Per-Class Metrics for MAPPO MoE (UTD-MHAD)}
\label{tab:confusion_matrix_utd}
\tiny
\resizebox{\textwidth}{!}{%
\begin{tabular}{lcccccccccccccccccccccccccccccc}
\hline
\textbf{True $\backslash$ Pred} & \textbf{Swipe l.} & \textbf{Swipe r.} & \textbf{Wave} & \textbf{Clap} & \textbf{Throw} & \textbf{Arm cro.} & \textbf{Basketb.} & \textbf{Draw X} & \textbf{Draw ci.} & \textbf{Draw ci.} & \textbf{Draw tr.} & \textbf{Bowling} & \textbf{Boxing} & \textbf{Basebal.} & \textbf{Tennis .} & \textbf{Arm curl} & \textbf{Tennis .} & \textbf{Push} & \textbf{Knock} & \textbf{Catch} & \textbf{Pickup .} & \textbf{Jog} & \textbf{Walk} & \textbf{Sit to .} & \textbf{Stand t.} & \textbf{Lunge} & \textbf{Squat} & \textbf{Prec.} & \textbf{Rec.} & \textbf{F1} \\
\hline
\textbf{Swipe l.} & 16 & 0 & 0 & 0 & 0 & 0 & 0 & 0 & 0 & 0 & 0 & 0 & 0 & 0 & 0 & 0 & 0 & 0 & 0 & 0 & 0 & 0 & 0 & 0 & 0 & 0 & 0 & 0.89 & 1.00 & 0.94 \\
\textbf{Swipe r.} & 0 & 16 & 0 & 0 & 0 & 0 & 0 & 0 & 0 & 0 & 0 & 0 & 0 & 0 & 0 & 0 & 0 & 0 & 0 & 0 & 0 & 0 & 0 & 0 & 0 & 0 & 0 & 0.89 & 1.00 & 0.94 \\
\textbf{Wave} & 0 & 0 & 16 & 0 & 0 & 0 & 0 & 0 & 0 & 0 & 0 & 0 & 0 & 0 & 0 & 0 & 0 & 0 & 0 & 0 & 0 & 0 & 0 & 0 & 0 & 0 & 0 & 0.94 & 1.00 & 0.97 \\
\textbf{Clap} & 0 & 0 & 0 & 16 & 0 & 0 & 0 & 0 & 0 & 0 & 0 & 0 & 0 & 0 & 0 & 0 & 0 & 0 & 0 & 0 & 0 & 0 & 0 & 0 & 0 & 0 & 0 & 0.84 & 1.00 & 0.91 \\
\textbf{Throw} & 0 & 0 & 0 & 0 & 14 & 0 & 0 & 1 & 0 & 0 & 0 & 0 & 0 & 0 & 0 & 0 & 0 & 1 & 0 & 0 & 0 & 0 & 0 & 0 & 0 & 0 & 0 & 1.00 & 0.88 & 0.93 \\
\textbf{Arm cro.} & 0 & 0 & 0 & 0 & 0 & 16 & 0 & 0 & 0 & 0 & 0 & 0 & 0 & 0 & 0 & 0 & 0 & 0 & 0 & 0 & 0 & 0 & 0 & 0 & 0 & 0 & 0 & 1.00 & 1.00 & 1.00 \\
\textbf{Basketb.} & 0 & 0 & 0 & 0 & 0 & 0 & 16 & 0 & 0 & 0 & 0 & 0 & 0 & 0 & 0 & 0 & 0 & 0 & 0 & 0 & 0 & 0 & 0 & 0 & 0 & 0 & 0 & 1.00 & 1.00 & 1.00 \\
\textbf{Draw X} & 0 & 0 & 0 & 0 & 0 & 0 & 0 & 16 & 0 & 0 & 0 & 0 & 0 & 0 & 0 & 0 & 0 & 0 & 0 & 0 & 0 & 0 & 0 & 0 & 0 & 0 & 0 & 0.89 & 1.00 & 0.94 \\
\textbf{Draw ci.} & 0 & 0 & 0 & 0 & 0 & 0 & 0 & 0 & 16 & 0 & 0 & 0 & 0 & 0 & 0 & 0 & 0 & 0 & 0 & 0 & 0 & 0 & 0 & 0 & 0 & 0 & 0 & 1.00 & 1.00 & 1.00 \\
\textbf{Draw ci.} & 0 & 0 & 0 & 0 & 0 & 0 & 0 & 0 & 0 & 16 & 0 & 0 & 0 & 0 & 0 & 0 & 0 & 0 & 0 & 0 & 0 & 0 & 0 & 0 & 0 & 0 & 0 & 0.80 & 1.00 & 0.89 \\
\textbf{Draw tr.} & 0 & 0 & 0 & 0 & 0 & 0 & 0 & 0 & 0 & 4 & 12 & 0 & 0 & 0 & 0 & 0 & 0 & 0 & 0 & 0 & 0 & 0 & 0 & 0 & 0 & 0 & 0 & 1.00 & 0.75 & 0.86 \\
\textbf{Bowling} & 0 & 0 & 0 & 0 & 0 & 0 & 0 & 0 & 0 & 0 & 0 & 16 & 0 & 0 & 0 & 0 & 0 & 0 & 0 & 0 & 0 & 0 & 0 & 0 & 0 & 0 & 0 & 1.00 & 1.00 & 1.00 \\
\textbf{Boxing} & 0 & 0 & 0 & 0 & 0 & 0 & 0 & 0 & 0 & 0 & 0 & 0 & 16 & 0 & 0 & 0 & 0 & 0 & 0 & 0 & 0 & 0 & 0 & 0 & 0 & 0 & 0 & 1.00 & 1.00 & 1.00 \\
\textbf{Basebal.} & 0 & 0 & 0 & 0 & 0 & 0 & 0 & 0 & 0 & 0 & 0 & 0 & 0 & 16 & 0 & 0 & 0 & 0 & 0 & 0 & 0 & 0 & 0 & 0 & 0 & 0 & 0 & 1.00 & 1.00 & 1.00 \\
\textbf{Tennis .} & 1 & 0 & 0 & 0 & 0 & 0 & 0 & 0 & 0 & 0 & 0 & 0 & 0 & 0 & 15 & 0 & 0 & 0 & 0 & 0 & 0 & 0 & 0 & 0 & 0 & 0 & 0 & 1.00 & 0.94 & 0.97 \\
\textbf{Arm curl} & 0 & 0 & 0 & 0 & 0 & 0 & 0 & 0 & 0 & 0 & 0 & 0 & 0 & 0 & 0 & 16 & 0 & 0 & 0 & 0 & 0 & 0 & 0 & 0 & 0 & 0 & 0 & 1.00 & 1.00 & 1.00 \\
\textbf{Tennis .} & 0 & 0 & 0 & 0 & 0 & 0 & 0 & 0 & 0 & 0 & 0 & 0 & 0 & 0 & 0 & 0 & 16 & 0 & 0 & 0 & 0 & 0 & 0 & 0 & 0 & 0 & 0 & 1.00 & 1.00 & 1.00 \\
\textbf{Push} & 0 & 0 & 0 & 0 & 0 & 0 & 0 & 0 & 0 & 0 & 0 & 0 & 0 & 0 & 0 & 0 & 0 & 16 & 0 & 0 & 0 & 0 & 0 & 0 & 0 & 0 & 0 & 0.84 & 1.00 & 0.91 \\
\textbf{Knock} & 0 & 0 & 1 & 3 & 0 & 0 & 0 & 0 & 0 & 0 & 0 & 0 & 0 & 0 & 0 & 0 & 0 & 0 & 12 & 0 & 0 & 0 & 0 & 0 & 0 & 0 & 0 & 0.92 & 0.75 & 0.83 \\
\textbf{Catch} & 1 & 2 & 0 & 0 & 0 & 0 & 0 & 1 & 0 & 0 & 0 & 0 & 0 & 0 & 0 & 0 & 0 & 2 & 1 & 9 & 0 & 0 & 0 & 0 & 0 & 0 & 0 & 1.00 & 0.56 & 0.72 \\
\textbf{Pickup .} & 0 & 0 & 0 & 0 & 0 & 0 & 0 & 0 & 0 & 0 & 0 & 0 & 0 & 0 & 0 & 0 & 0 & 0 & 0 & 0 & 16 & 0 & 0 & 0 & 0 & 0 & 0 & 1.00 & 1.00 & 1.00 \\
\textbf{Jog} & 0 & 0 & 0 & 0 & 0 & 0 & 0 & 0 & 0 & 0 & 0 & 0 & 0 & 0 & 0 & 0 & 0 & 0 & 0 & 0 & 0 & 16 & 0 & 0 & 0 & 0 & 0 & 1.00 & 1.00 & 1.00 \\
\textbf{Walk} & 0 & 0 & 0 & 0 & 0 & 0 & 0 & 0 & 0 & 0 & 0 & 0 & 0 & 0 & 0 & 0 & 0 & 0 & 0 & 0 & 0 & 0 & 15 & 0 & 0 & 0 & 0 & 1.00 & 1.00 & 1.00 \\
\textbf{Sit to .} & 0 & 0 & 0 & 0 & 0 & 0 & 0 & 0 & 0 & 0 & 0 & 0 & 0 & 0 & 0 & 0 & 0 & 0 & 0 & 0 & 0 & 0 & 0 & 16 & 0 & 0 & 0 & 1.00 & 1.00 & 1.00 \\
\textbf{Stand t.} & 0 & 0 & 0 & 0 & 0 & 0 & 0 & 0 & 0 & 0 & 0 & 0 & 0 & 0 & 0 & 0 & 0 & 0 & 0 & 0 & 0 & 0 & 0 & 0 & 16 & 0 & 0 & 1.00 & 1.00 & 1.00 \\
\textbf{Lunge} & 0 & 0 & 0 & 0 & 0 & 0 & 0 & 0 & 0 & 0 & 0 & 0 & 0 & 0 & 0 & 0 & 0 & 0 & 0 & 0 & 0 & 0 & 0 & 0 & 0 & 16 & 0 & 1.00 & 1.00 & 1.00 \\
\textbf{Squat} & 0 & 0 & 0 & 0 & 0 & 0 & 0 & 0 & 0 & 0 & 0 & 0 & 0 & 0 & 0 & 0 & 0 & 0 & 0 & 0 & 0 & 0 & 0 & 0 & 0 & 0 & 15 & 1.00 & 1.00 & 1.00 \\
\hline
\end{tabular}%
}
\end{sidewaystable*}

Table \ref{tab:confusion_matrix_utd} reports the confusion matrix and per class precision, recall, and F1 score for MAPPO MoE on UTD MHAD across all 27 actions. Of the 430 test samples, 412 are correctly classified, giving an overall accuracy of 95.81 percent, with 22 of the 27 classes achieving perfect precision, recall, and F1 score. The remaining errors are concentrated in a small number of classes rather than spread across the full action set. Catch accounts for the largest share of errors, with predictions scattered across five different classes including Swipe left, Swipe right, Draw X, Push, and Knock, indicating that this action lacks a consistent, distinguishing signal captured by the model. Draw triangle is confused entirely with Draw circle counter clockwise, and Knock is confused mostly with Clap, both plausibly explained by the kinematic similarity between each pair of actions. These results indicate that the model performs reliably across most of the action set, with a small number of visually or kinematically similar actions accounting for nearly all of the remaining classification errors.

\begin{figure} 
    \centering
    \includegraphics[scale = 0.35]{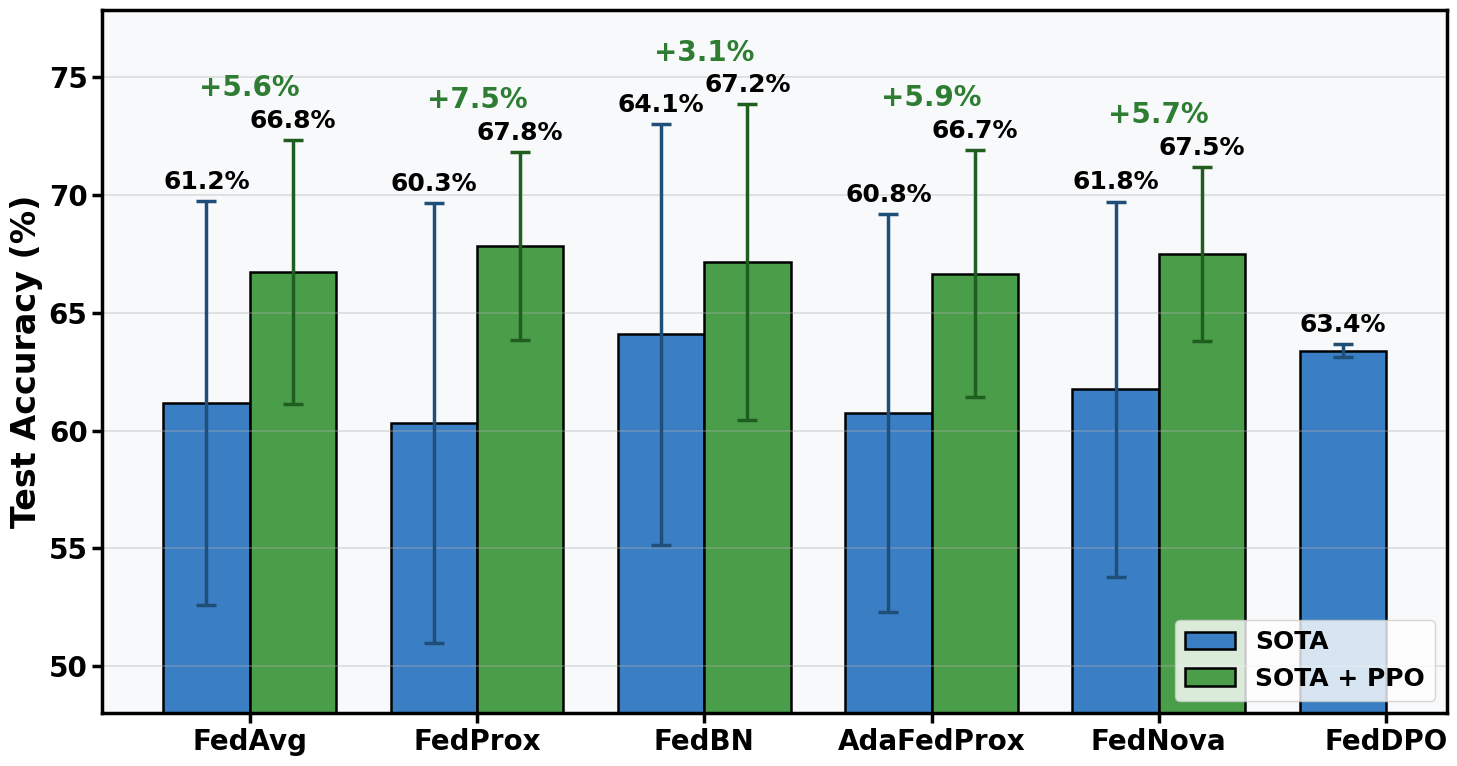}
    \caption{Test Accuracy: SOTA FL vs. Server PPO Aggregator for MEx Dataset}
    \label{fig:vanilla_MEx}
    \vspace{-3mm}
\end{figure}

Fig. \ref{fig:vanilla_MEx} compares the mean test accuracy of five vanilla FL baselines and their Server PPO-enhanced variants on the MEx physiotherapy exercise dataset. Results are averaged over five independent random seeds {42, 7, 123, 999, 2024}, using 30 communication rounds and 5 PPO warmup rounds. Error bars represent one standard deviation across seeds. The Server PPO aggregator consistently improves performance, achieving gains of +2.50\% (FedAvg), +2.92\% (FedProx), +6.25\% (FedBN), +6.67\% (AdaFedProx), and +0.83\% (FedNova). In addition, the PPO-enhanced variants generally exhibit lower variability across seeds, indicating improved training stability and reproducibility. Overall, the results demonstrate that adaptive PPO-based aggregation provides a consistent advantage over conventional static aggregation methods.

\subsubsection{\textit{UTD MHAD} Dataset}

\begin{figure*}[ht!]
\centering

\subfigure[]{
\includegraphics[width=0.47\linewidth]{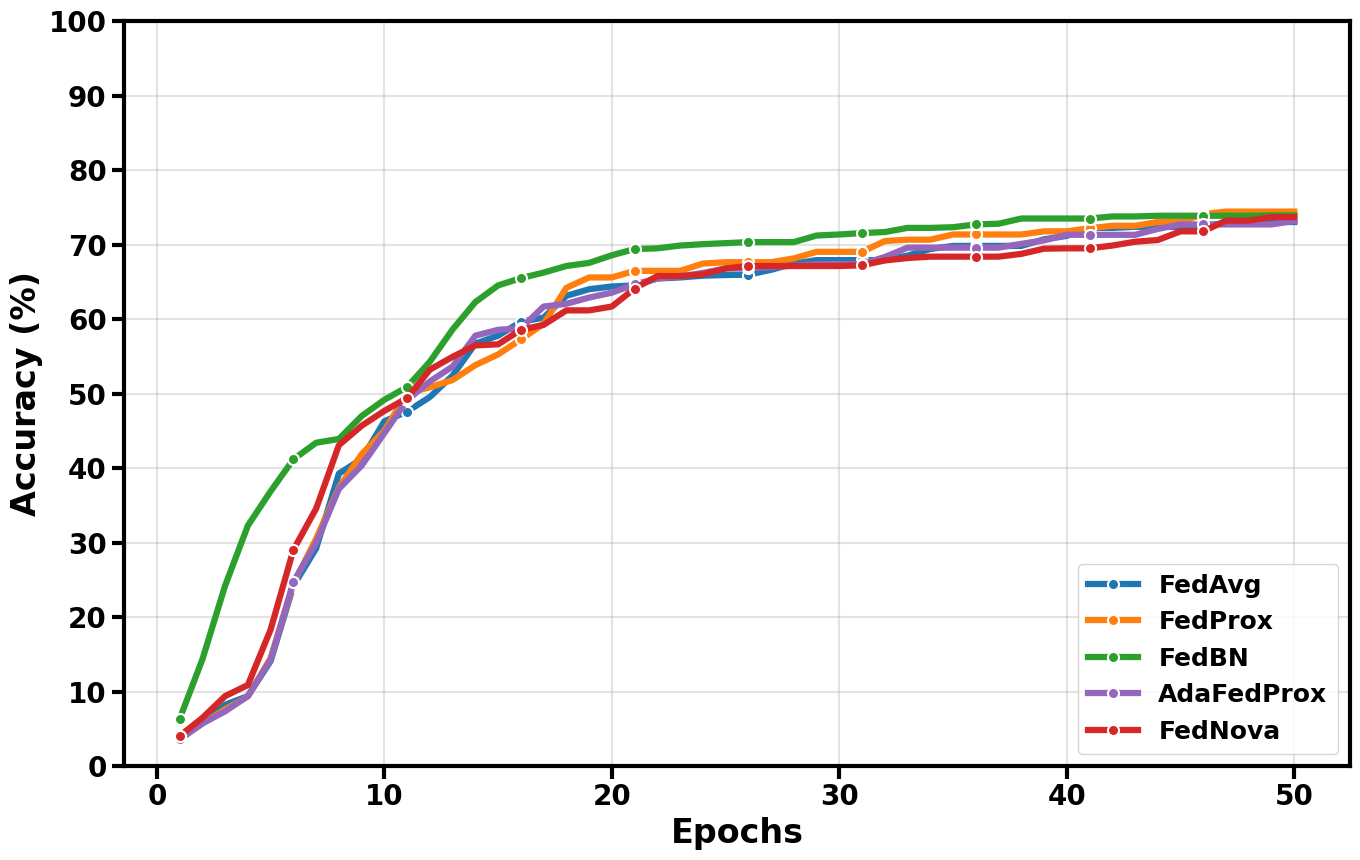}
}
\hfill
\subfigure[\label{kl_mqtt_len}]{
\includegraphics[width=0.47\linewidth]{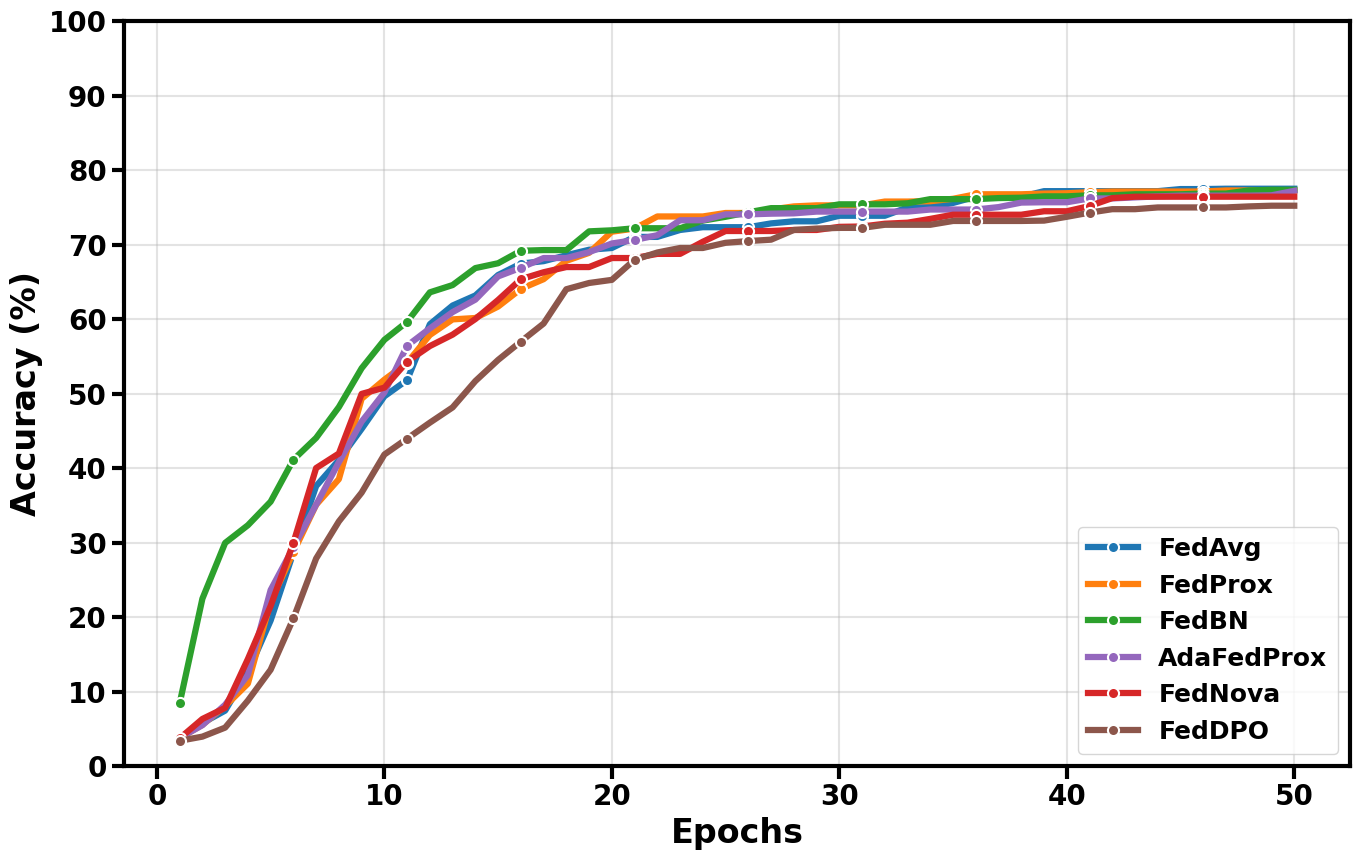}
}

\caption{(a) Test Accuracy of SOTA FL on UTD MHAD dataset; (b) Test Accuracy of SOTA FL using PPO on UTD MHAD dataset}
\label{fig:fl_utd}

\vspace{-3mm}
\end{figure*}
Fig. \ref{fig:fl_utd} presents the per-round test accuracy trajectory of all federated learning methods over 50 communication rounds on the UTD-MHAD dataset, which comprises 27 action classes captured from three sensor modalities, skeleton, inertial, and joint velocity (derived).

The proposed architecture exhibits significantly faster convergence than all baseline methods, reaching approximately $82\%$ accuracy by round 2 and exceeding $91\%$ by round 5, whereas the baselines remain between $35--60\%$ during the same period. The proposed method also achieves the highest peak accuracy of $96.05\%$, outperforming FedAvg $(95.81\%)$, AdaFedProx $(92.79\%)$, FedNova $(91.63\%)$, and FedDPO $(91.16\%)$. Moreover, it maintains a stable accuracy range of $93-96\%$ throughout training, while the baselines exhibit greater fluctuations and slower convergence.

The superior performance of FL PPO FedAvg stems from its ability to dynamically learn modality-specific aggregation weights through PPO agents. By jointly optimizing classification performance and sensor cost, the agents quickly identify the most informative modalities and assign aggregation weights accordingly. This adaptive aggregation strategy is particularly effective for UTD-MHAD, where modalities differ significantly in their information content and dimensionality. Consequently, the proposed method achieves faster convergence, improved stability, and consistently higher accuracy than conventional federated aggregation approaches.




\begin{figure} 
    \centering
    \includegraphics[scale = 0.35]{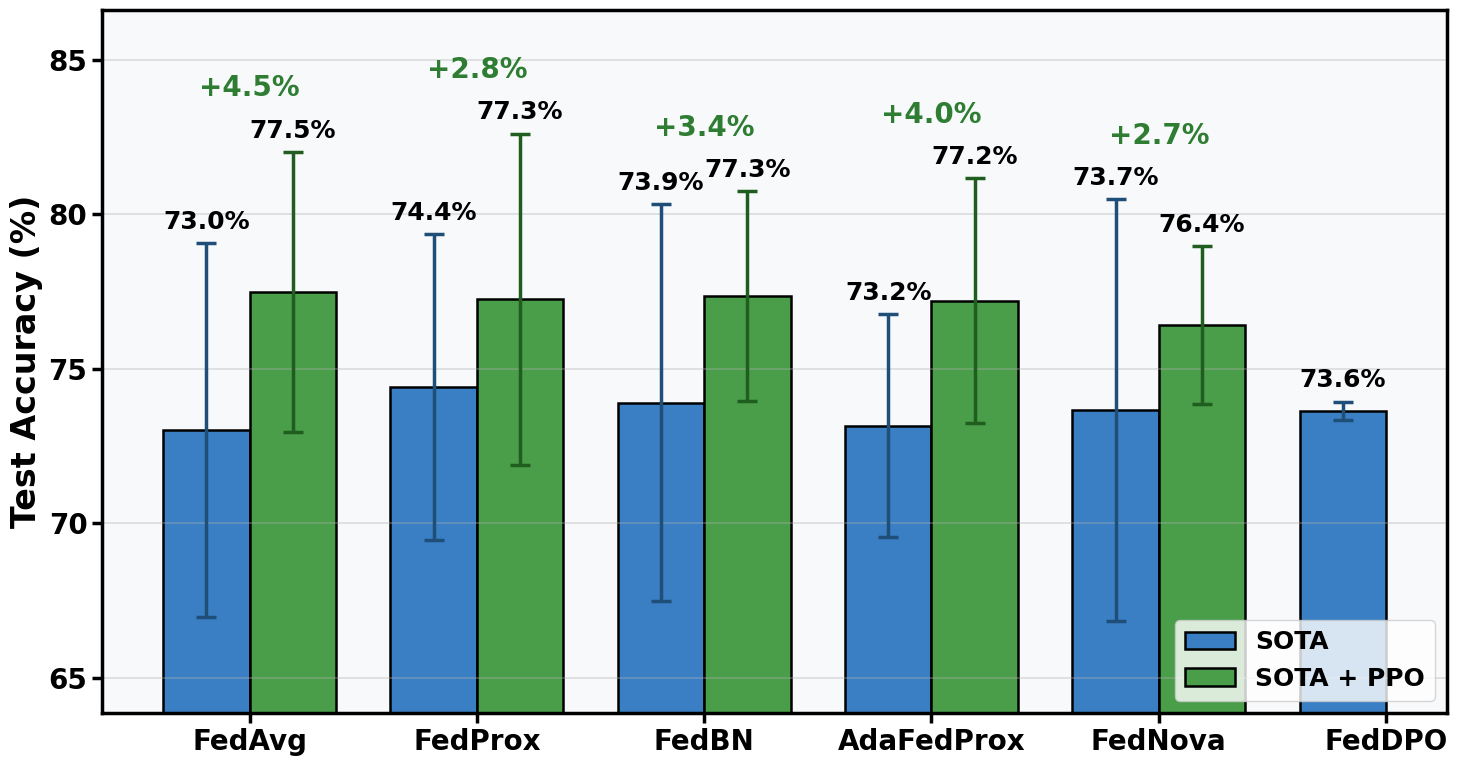}
    \caption{Test Accuracy: SOTA FL vs. Server PPO Aggregator for \textit{UTD MHAD} Dataset}
    \label{fig:vanilla_utdmhad}
    \vspace{-3mm}
\end{figure}

Fig. \ref{fig:vanilla_utdmhad} presents the UTD-MHAD 5-seed ablation study comparing vanilla FL baselines and their Server PPO-enhanced variants across five random seeds {42, 7, 123, 999, 2024}, evaluated over 50 communication rounds with 5 warmup rounds and 5 local epochs per round. The Server PPO aggregator maintains or slightly improves performance for FedAvg, FedProx, FedBN, and AdaFedProx, indicating that these methods already achieve near-optimal performance on the dataset. The largest gain is observed for FedNova, where Server PPO improves the mean accuracy from 90.19\% to 91.02\% (+0.84\%) while reducing the standard deviation from $\pm$1.82\% to $\pm$1.30\%. These results demonstrate that adaptive PPO-based aggregation can improve both accuracy and robustness, particularly for methods sensitive to variations in client updates.

\begin{table}[H]
\centering
\caption{5-Seed Ablation Study: SOTA FL vs. Server PPO Aggregator for \textit{UTD MHAD} Dataset}
\label{tab:utd_ablation_5seed}
\setlength{\tabcolsep}{8pt}
\renewcommand{\arraystretch}{1.3}
\begin{tabular}{l c c}
\toprule
\textbf{Method}
& \textbf{Vanilla Mean$\pm$Std}
& \textbf{ServerPPO Mean$\pm$Std} \\
\midrule
FedAvg      & 73.0 $\pm$ 4.43 & \textbf{77.5 $\pm$ 3.26} \\
FedProx     & 74.40 $\pm$ 2.87 & \textbf{77.30 $\pm$ 1.59} \\
FedBN       & 73.9 $\pm$ 2.65 & \textbf{77.3 $\pm$ 2.95} \\
AdaFedProx  & 73.26 $\pm$ 2.62 & \textbf{77.2 $\pm$ 1.84} \\
FedNova     & 73.7 $\pm$ 3.55 & \textbf{76.4 $\pm$ 2.27} \\
 FedDPO      & 73.6 $\pm$ 2.55 & -- \\
\bottomrule
\end{tabular}
\end{table}

Table~\ref{tab:utd_ablation_5seed} presents the per-seed results of the UTD-MHAD 5-seed ablation study across seeds {42, 7, 123, 999, 2024}. Overall, Server PPO achieves higher or equal accuracy in 18 out of 25 seed-method comparisons, corresponding to a non-deterioration rate of 72\%. The most consistent gains are observed for FedNova, where Server PPO improves performance in four out of five seeds, increasing the mean accuracy from 90.19\% to 91.02\% (+0.84\%) while reducing the standard deviation from $\pm$1.82\% to $\pm$1.30\%. In contrast, FedAvg, FedProx, FedBN, and AdaFedProx exhibit only marginal changes (within $\pm$0.05\%), suggesting that their aggregation dynamics are already near convergence. These results indicate that the Server PPO policy is particularly effective when meaningful inter-client variability exists, enabling more stable and accurate aggregation than static weighting schemes.

\section{Baseline Results}
This section presents the comparison of the performance of \textit{FedMHAR} with respect to the state-of-the-art federated methods.

\subsection{MEx Dataset}

\begin{table}[H]
\centering
\caption{Comparison with Baselines on MEx}
\label{tab:comparison_baselines}
\setlength{\tabcolsep}{5pt}
\renewcommand{\arraystretch}{1.25}
\begin{tabular}{l l c}
\toprule
\textbf{Category} & \textbf{Method} & \textbf{Accuracy (\%)} \\
\midrule
Prior Baselines & LSTM-CNN~\cite{xia2020lstm} & $84.61 \pm 3.09$ \\
 & MhaGNN~\cite{wang2023mhagnn} & $87.11 \pm 2.62$ \\
 & TCN-Inception~\cite{al2024tcn} & $81.38 \pm 3.22$ \\
 & HART~\cite{ek2023transformer} & $80.00 \pm 4.35$ \\
 & MobileHART~\cite{ek2023transformer} & $84.08 \pm 2.74$ \\
\midrule
Centralised & Baseline MoE & $84.80 \pm 2.53$ \\
 & \textbf{MAPPO-MoE} & \textbf{$87.11 \pm 2.83$} \\
\bottomrule
\end{tabular}
\end{table}

Table~\ref{tab:comparison_baselines} compares MAPPO-MoE with LSTM-CNN, MhaGNN, and TCN-Inception on the MEx dataset. While TCN-Inception and MhaGNN benefit from multi-scale temporal modelling and graph based cross modal learning, respectively, MAPPO-MoE achieves the highest accuracy of $87.30\%\pm2.53\%$. The improvement stems from its PPO-based adaptive fusion policy, which dynamically adjusts modality contributions on a per sample basis while considering sensor cost. The lower standard deviation further indicates more stable performance across random seeds.

\subsection{\textit{UTD MHAD}}

\begin{table}[H]
\centering
\caption{Comparison with Baselines on \textit{UTD-MHAD}}
\label{tab:utd_comparison_baselines}
\setlength{\tabcolsep}{5pt}
\renewcommand{\arraystretch}{1.25}
\begin{tabular}{l l c}
\toprule
\textbf{Category} & \textbf{Method} & \textbf{Accuracy (\%)} \\
\midrule
Prior Baselines & LSTM-CNN~\cite{xia2020lstm} & $92.74 \pm 0.98$ \\
 & MhaGNN~\cite{wang2023mhagnn} & $92.88 \pm 0.54$ \\
 & TCN-Inception~\cite{al2024tcn} & $93.40 \pm 0.71$ \\
 & HART~\cite{ek2023transformer} & $61.77 \pm 2.61$ \\ 
 & MobileHART~\cite{ek2023transformer} & $75.07 \pm 2.53$ \\
\midrule
Centralised & Baseline MoE & $82.84 \pm 0.67$ \\
 & \textbf{MAPPO-MoE} & \textbf{$94.98 \pm 0.88$} \\
\midrule
\bottomrule
\end{tabular}
\end{table}

Table~\ref{tab:utd_comparison_baselines} presents results on the more heterogeneous UTD-MHAD dataset. MAPPO-MoE achieves $94.98\%\pm0.88\%$, outperforming all baselines by a significant margin. The gain highlights the effectiveness of adaptive modality weighting when sensor streams provide complementary information. Unlike fixed-weight or static graph-based fusion methods, MAPPO-MoE learns activity-dependent modality importance, leading to both higher accuracy and the lowest variance across seeds, demonstrating robust and consistent performance.

\section{Ablation Study} \label{sec:ablation}

Table~\ref{tab:MEx_modality_ablation} presents the centralized modality ablation results on \textit{MEx}, comparing the baseline MoE with the proposed MAPPO-guided MoE across all single-, double-, triple-, and full-modality subsets (3-seed mean$\pm$std). The results show that the benefit of MAPPO increases with the number of available modalities. Using all four modalities, MAPPO improves accuracy from 0.84 to 0.91 while reducing the standard deviation from $\pm$0.04 to $\pm$0.01, demonstrating both higher accuracy and improved stability.

For single-modality settings, MAPPO provides little benefit, as there is no expert coordination to optimize. It yields marginal gains for \texttt{act} and \texttt{pm}, but slightly degrades performance for \texttt{acw} and \texttt{dc}. In contrast, consistent improvements are observed in multi-modality settings where adaptive expert weighting becomes important. The largest gains are achieved for complementary modality pairs such as \texttt{act+pm} (+0.06) and for challenging three-modality subsets, particularly when \texttt{dc} is removed (+0.09).

Overall, the results indicate that MAPPO is most effective in complex multimodal scenarios where expert coordination is essential, while remaining largely neutral when fusion offers limited benefit. This demonstrates that the proposed policy learns meaningful modality interactions and provides the greatest advantage in practical multimodal HAR settings.

\begin{table}[H]
\centering
\caption{Centralized Modality Ablation on \textit{MEx} Dataset}
\label{tab:MEx_modality_ablation}
\setlength{\tabcolsep}{6pt}
\renewcommand{\arraystretch}{1.25}
\begin{tabular}{l c c}
\toprule
\textbf{Modalities} & \textbf{Baseline MoE (\%)} & \textbf{MAPPO MoE (\%)} \\
\midrule
acw      & \textbf{41.01} $\pm$ 0.76 & 35.20 $\pm$ 2.16 \\
act      & 61.95 $\pm$ 1.98 & \textbf{62.39 $\pm$ 1.00} \\
dc       & \textbf{77.74 $\pm$ 2.42} & 75.99 $\pm$ 2.16 \\
pm       & 66.78 $\pm$ 4.30 & \textbf{66.89 $\pm$ 6.65} \\
\midrule
acw+act  & 68.31 $\pm$ 2.38 & \textbf{71.16 $\pm$ 1.00} \\
acw+dc   & 76.21 $\pm$ 2.01 & 75.33 $\pm$ 0.33 \\
acw+pm   & 64.04 $\pm$ 1.87 & 60.86 $\pm$ 3.56 \\
act+dc   & 78.51 $\pm$ 2.84 & \textbf{78.95 $\pm$ 1.74} \\
act+pm   & 68.20 $\pm$ 2.66 & \textbf{74.67 $\pm$ 0.57} \\
dc+pm    & 83.11 $\pm$ 3.13 & \textbf{79.21 $\pm$ 4.61} \\
\midrule
acw+act+dc  & 76.10 $\pm$ 1.25 & \textbf{81.58 $\pm$ 3.43} \\
acw+act+pm  & 65.57 $\pm$ 1.00 & \textbf{75.00 $\pm$ 6.03} \\
acw+pm+dc & \textbf{82.35 $\pm$ 0.38} & 82.13 $\pm$ 2.14 \\
act+pm+dc & 84.76 $\pm$ 2.24 & \textbf{86.51 $\pm$ 2.30} \\
\midrule
ALL       & 80.43 $\pm$ 4.19 & \textbf{87.79 $\pm$ 0.57} \\
\bottomrule
\end{tabular}
\end{table}

\begin{table}[H]
\centering
\caption{Centralized Modality Ablation on \textit{UTD-MHAD} (5-seed).}
\label{tab:utd_modality_ablation}
\setlength{\tabcolsep}{6pt}
\renewcommand{\arraystretch}{1.25}
\begin{tabular}{l c c}
\toprule
\textbf{Modalities} & \textbf{Baseline MoE (\%)} & \textbf{MAPPO MoE (\%)} \\
\midrule
skeleton & 75.67 $\pm$ 0.58 & \textbf{76.00 $\pm$ 0.33} \\
inertial & 86.28 $\pm$ 0.71 & \textbf{87.21 $\pm$ 0.93} \\
joint\_vel & \textbf{71.67 $\pm$ 0.62} & 70.84 $\pm$ 0.84 \\
\midrule
inertial + skeleton & 91.91 $\pm$ 0.86 & \textbf{94.19 $\pm$ 2.28} \\
skeleton + joint\_vel & 80.88 $\pm$ 0.79 & \textbf{81.91 $\pm$ 1.02} \\
inertial + joint\_vel & 91.81 $\pm$ 0.91 & \textbf{93.21 $\pm$ 1.40} \\
\midrule
ALL & 93.40 $\pm$ 0.74 & \textbf{95.35 $\pm$ 1.95} \\
\bottomrule
\end{tabular}
\end{table}

Table~\ref{tab:utd_modality_ablation} presents the 5-seed centralized modality ablation on UTD-MHAD. Among individual modalities, the inertial sensor achieves the highest accuracy ($86.28\% \pm 0.71$), outperforming skeleton ($75.67\% \pm 0.58$) and joint velocity ($71.67\% \pm 0.62$). Combining modalities consistently improves performance, with the inertial+skeleton pair reaching $91.91\% \pm 0.86$, while using all three modalities achieves the highest Baseline MoE accuracy of $93.40\% \pm 0.74$. MAPPO further improves performance in six of the seven settings, with the largest gains observed for inertial+skeleton ($91.91\%\rightarrow94.19\%$, +2.28\%) and the full-modality configuration ($93.40\%\rightarrow95.35\%$, +1.95\%). The only performance drop occurs for the joint velocity-only setting, where cross-modal coordination is absent. Overall, these results demonstrate that MAPPO is most effective in multi-modal scenarios, where adaptive expert weighting can better exploit complementary information, albeit with a slight increase in variance across seeds.

\section{Error Analysis}


\begin{figure} 
    \centering
    \includegraphics[scale = 0.35]{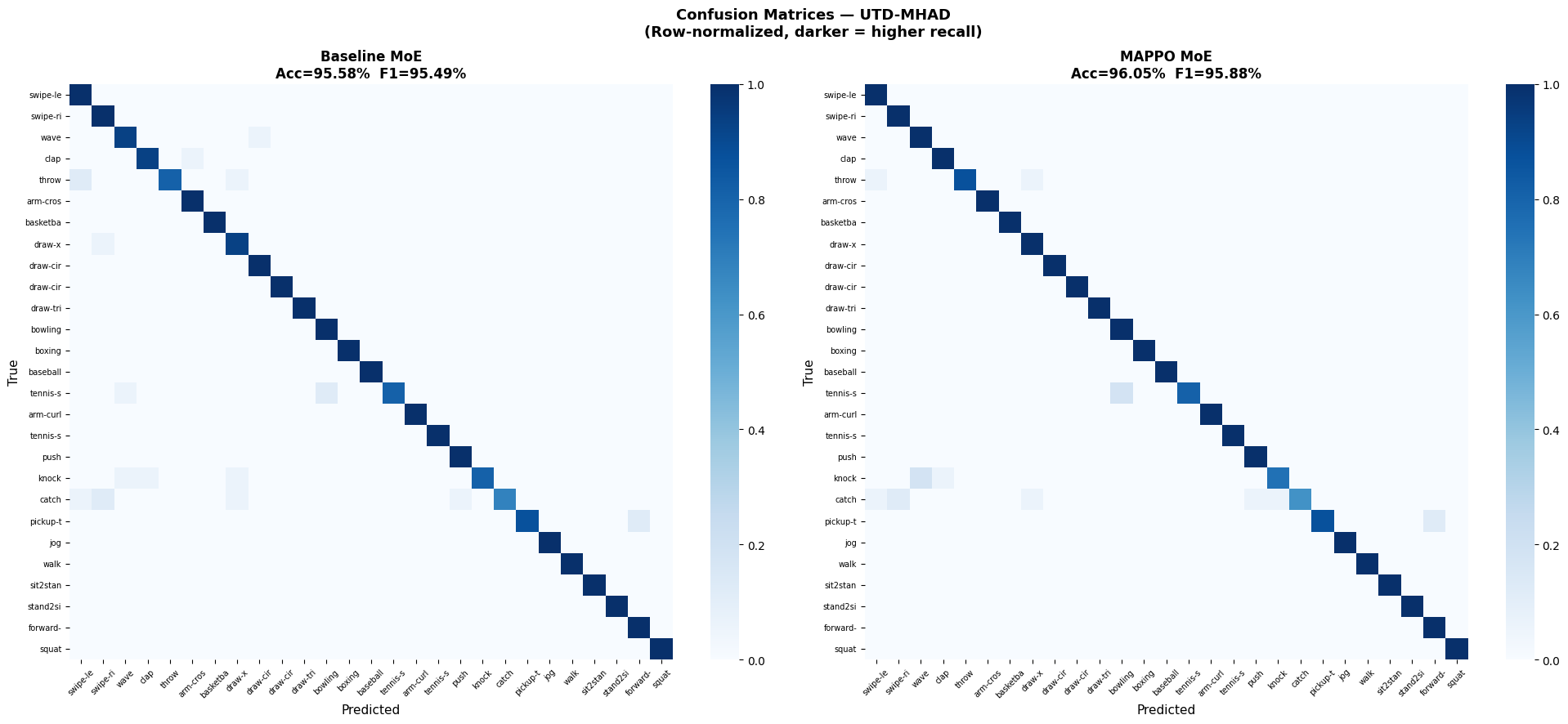}
    \caption{Confusion matrices of Baseline MoE and MAPPO-MoE on the UTD MHAD Dataset}
    \label{fig:con_utd}
\end{figure}

\begin{figure} 
    \centering
    \includegraphics[scale = 0.35]{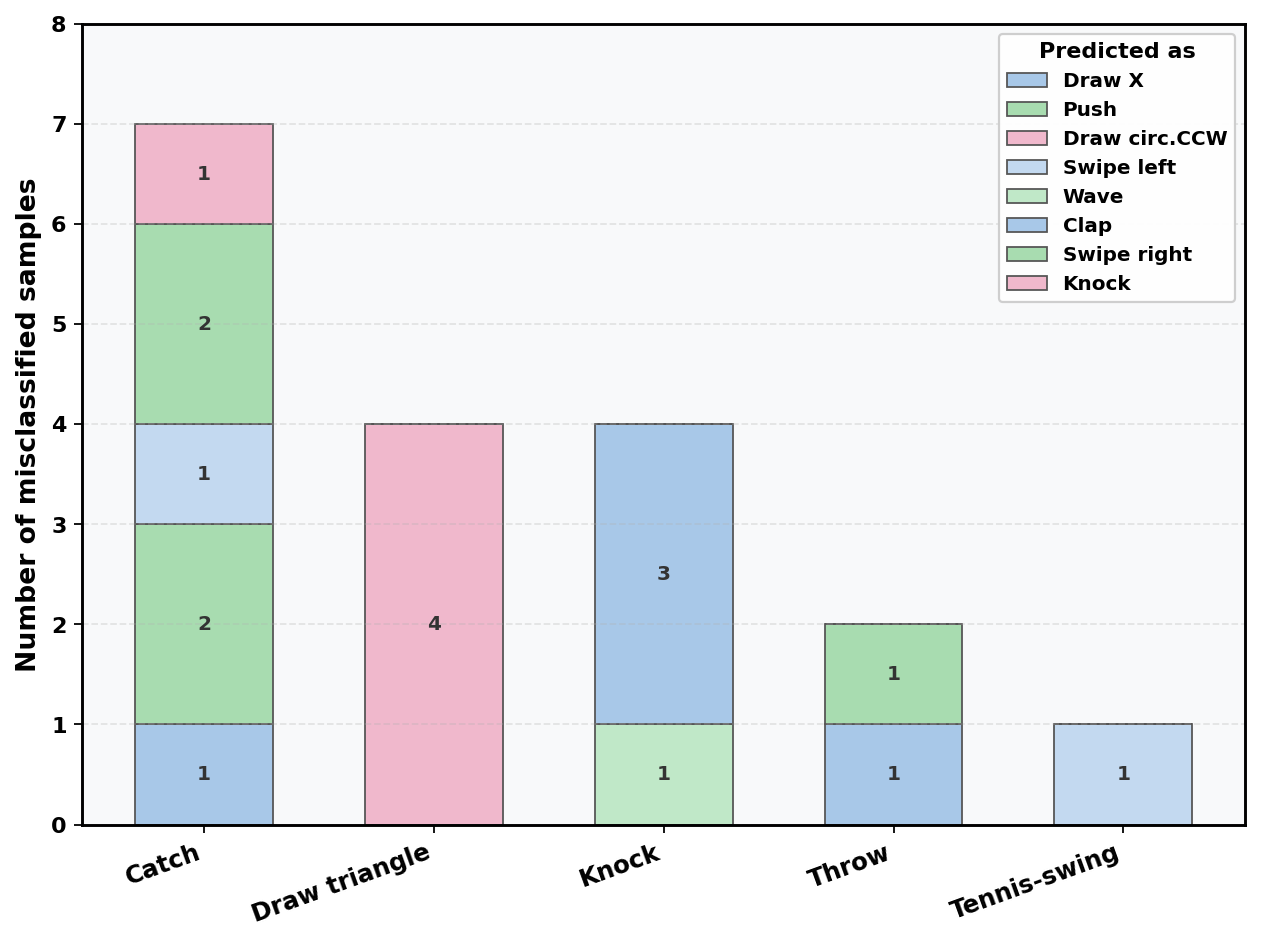}
    \caption{Confusion matrices of Baseline MoE and MAPPO-MoE on the UTD MHAD Dataset}
    \label{fig:misclass_breakdown}
\end{figure}

Fig. \ref{fig:con_utd} represents the normalized confusion matrices of Baseline MoE and MAPPO-MoE on the UTD MHAD dataset. The confusion matrices highlight the class-wise recognition performance of the two models and the impact of adaptive modality fusion on reducing inter-class confusion. Both models achieve high recognition accuracy on UTD-MHAD, with most activities concentrated along the diagonal, indicating strong classification performance. The remaining errors are primarily confined to kinematically similar actions such as catch, knock, throw, bowling, and tennis swing, where subtle motion differences make discrimination more challenging. The proposed MAPPO-MoE slightly reduces these residual confusions, particularly for sports-related actions, resulting in improvements in both accuracy (95.58\% to 96.05\%) and weighted F1-score (95.49\% to 95.88\%). The sharper diagonal structure of the MAPPO-MoE confusion matrix suggests that adaptive modality weighting produces more discriminative feature representations, leading to more reliable recognition of challenging activities.

Fig.~\ref{fig:misclass_breakdown} shows the $18$ total misclassifications, broken down by true class and predicted class. \textit{Catch} accounts for the most errors ($7$ of $18$), spread across five different predicted classes. \textit{Draw triangle}'s $4$ errors are entirely confused with \textit{Draw circle (CCW)}, and \textit{Knock}'s $4$ errors are mostly confused with \textit{Clap} ($3$ of $4$), both plausibly due to kinematic similarity between the action pairs. \textit{Throw} ($2$ errors) and \textit{Tennis-swing} ($1$ error) contribute minor, isolated confusions. This concentrated, explainable error pattern is consistent with the model's high overall accuracy and F1-score ($95.81\%$ and $95.88\%$).

\section{Conclusion and Future Work}

In this work, we presented a two-part framework for cost-aware, adaptive multimodal fusion in human activity recognition, addressing a limitation shared by conventional fusion methods: the assumption that all sensor modalities carry equal, time-invariant importance, regardless of their acquisition cost or per-sample reliability. In the centralized setting, we formulated multimodal fusion as a cooperative multi-agent reinforcement learning problem, in which a PPO-based agent per modality learns a per-sample fusion weight, enabling the Mixture-of-Experts network to dynamically emphasize informative sensors while down-weighting costlier ones whenever cheaper modalities already provide sufficient information. Extending this idea to the federated setting, we introduced BiFL-PPO, a bidirectional federated optimization strategy in which a server-side PPO policy learns client-specific trust weights that are fed back to clients to jointly guide server-side aggregation and client-side local optimization, rather than treating these as independent decisions.

Evaluated on the MEx Rehabilitation and UTD-MHAD datasets, our centralized MARL-based framework achieved $87.30\%$ and $94.98\%$ classification accuracy, respectively, outperforming conventional fixed-weight fusion and existing HAR baselines, while our federated FedMHAR framework achieved $79.74\%$ and $77.49\%$ accuracy, consistently surpassing FedAvg, FedProx, FedBN, FedNova, and AdaFedProx. Beyond raw accuracy, our federated results show that the proposed bidirectional trust-weighting strategy provides a consistent, positive gain when layered on top of every one of these baseline methods individually, rather than acting as a replacement for their existing heterogeneity-handling mechanisms, indicating that adaptive, learned aggregation and each method's own local optimization strategy (proximal regularization, update normalization, local batch-normalization) address complementary aspects of federated heterogeneity rather than competing solutions to the same problem.

These results support our central claim: treating sensor trust and client trust as learned, context-dependent decisions, rather than fixed a priori assumptions, yields consistent improvements in both accuracy and training stability across two datasets, two learning paradigms (centralized and federated), and five distinct federated baselines, while additionally reducing reliance on costlier sensor modalities where cheaper alternatives suffice. We view this as evidence that reinforcement-learning-based trust and cost modeling is a viable, general-purpose mechanism for multimodal and federated HAR systems more broadly, rather than one narrowly tuned to a specific fusion architecture or aggregation rule.

Limitations of this work point toward natural directions for future research. Our centralized modality ablation experiments (Section~\ref{sec:ablation}) show that learned weighting provides the clearest benefit when there is genuine, exploitable heterogeneity across modalities or clients to detect; evaluating our framework under deliberately induced sensor degradation or corruption, rather than clean, well-controlled inputs, would more directly test the cost-aware and reliability-aware aspects of our design. Additionally, while we evaluate against five representative federated baselines, extending this comparison to include recently proposed reinforcement-learning-based federated methods, and validating BiFL-PPO's per-round reward design under a larger number of clients and communication rounds, remain important directions for establishing the broader generality of our approach. Finally, extending the framework to longitudinal monitoring in clinical populations is a natural next step. For example, six months of continuous smart-ring monitoring in patients with cirrhosis showed that sleep regularity, a property observable only across nights, was the strongest sleep-related correlate of health-related quality of life \cite{buckholz2026sleep}. Such settings, which combine inexpensive passive wearable streams with costlier clinical assessments across multiple sites, are a natural application for cost-aware and federated multimodal fusion.

\section*{Conflict of interest statement} There is no conflict of interest among the authors of this manuscript.

\section*{Data availability statement}
The datasets used in this study, namely the MEx and UTD-MHAD datasets, are publicly available open-source datasets and can be accessed from their respective original sources.

\section*{Funding}
The research work of Debasmita Dey is supported by the Anusandhan National Research Foundation (ANRF), India, through the National Post-Doctoral Fellowship (NPDF) project grant, administered by the Indian Statistical Institute, Kolkata (Project No. \textit{PDF/2026/002233}).

\bibliographystyle{unsrt}
\bibliography{biblio}

\end{document}